\documentclass{article}

\usepackage{PRIMEarxiv}

\usepackage[utf8]{inputenc} 
\usepackage[T1]{fontenc}    
\usepackage{url}            
\usepackage{booktabs}       
\usepackage{amsfonts}       
\usepackage{nicefrac}       
\usepackage{microtype}      
\usepackage{lipsum}
\usepackage{fancyhdr}       
\usepackage{graphicx}       
\usepackage{amsmath}
\usepackage{multirow}
\usepackage{subcaption}
\usepackage{dsfont}
\usepackage{url}

\graphicspath{{media/}}     

\usepackage{bm}
\usepackage{adjustbox}

\usepackage{algorithm}
\usepackage{algorithmic}
\usepackage{amsthm}

\newtheorem{defn}{Definition}[section]

\usepackage{newfloat}
\usepackage{listings}
\floatstyle{ruled}
\newfloat{listing}{tb}{lst}{}
\floatname{listing}{Listing}
\usepackage{bibentry}
\usepackage{xcolor}

\usepackage{amsfonts}
\usepackage{graphicx}
\usepackage{adjustbox}
\usepackage{amssymb}
\usepackage{pifont}
\usepackage{multirow}
\usepackage{amsmath}
\usepackage{bbm}
\usepackage{float}
\usepackage{caption}
\usepackage{subcaption}
\usepackage{dsfont}
\usepackage{etoolbox}
\usepackage{chronosys}
\usepackage{lscape}
\usepackage{longtable}
\usepackage{amsmath}
\usepackage{nccmath}
\usepackage{xcolor}

\title{Knowledge Graph Embedding: An Overview}

\author{
  Xiou Ge \\
  University of Southern California \\
  Los Angeles, USA\\
   \And
  Yun-Cheng Wang \\
  University of Southern California \\
  Los Angeles, USA\\
   \And
  Bin Wang \\
  Institute for Infocomm Research (I2R) \\
  A*STAR \\
  Singapore\\
   \And
  C.-C. Jay Kuo \\
  University of Southern California \\
  Los Angeles, USA\\
}

\begin{document}
\maketitle

\begin{abstract}
Many mathematical models have been leveraged to design embeddings for representing Knowledge Graph (KG) entities and relations for link prediction and many downstream tasks. These mathematically-inspired models are not only highly scalable for inference in large KGs, but also have many explainable advantages in modeling different relation patterns that can be validated through both formal proofs and empirical results.  In this paper, we make a comprehensive overview of the current state of research in KG completion. In particular, we focus on two main branches of KG embedding (KGE) design: 1) distance-based methods and 2) semantic matching-based methods. We discover the connections between recently proposed models and present an underlying trend that might help researchers invent novel and more effective models. Next, we delve into CompoundE and CompoundE3D, which draw inspiration from 2D and 3D affine operations, respectively. They encompass a broad spectrum of techniques including distance-based and semantic-based methods. We will also discuss an emerging approach for KG completion which leverages pre-trained language models (PLMs) and textual descriptions of entities and relations and offer insights into the integration of KGE embedding methods with PLMs for KG completion.
\end{abstract}


\renewcommand{\figurename}{Figure}
\renewcommand{\tablename}{Table}

\section{Introduction}\label{intro}

\begin{figure}[ht]
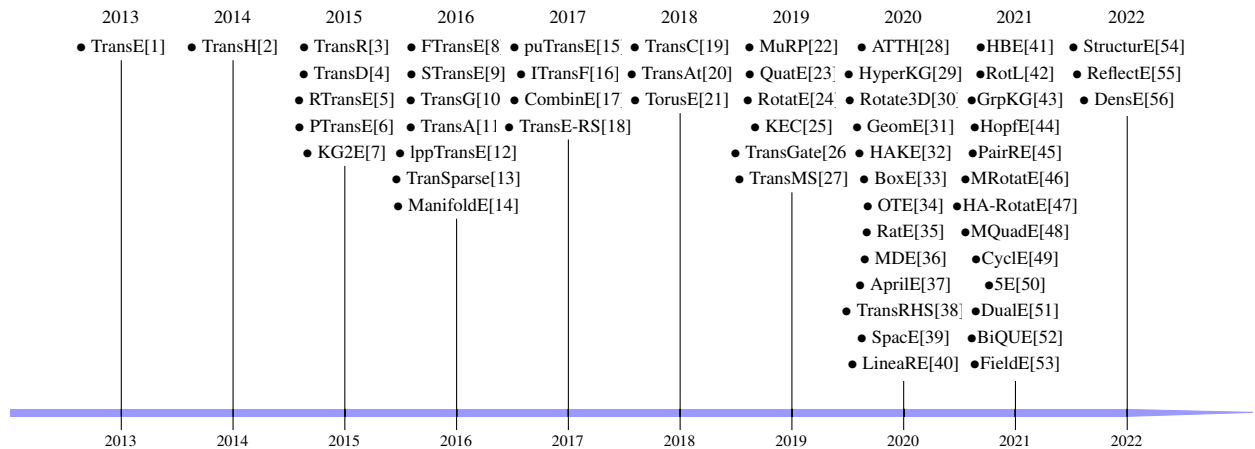

\vspace{-150pt}
\startchronology[startyear=2012,stopyear=2022, startdate=false, color=blue!40, stopdate=false, arrow=true, height=3pt]
\setupchronoevent{textstyle=\scriptsize,datestyle=\scriptsize}
\chronoevent[markdepth=-140pt]{2013}{$\bullet$ TransE\cite{bordes2013translating}}

\chronoevent[markdepth=-140pt]{2014}{$\bullet$ TransH\cite{wang2014knowledge}}

\chronoevent[markdepth=-140pt,mark=false]{2015}{$\bullet$ TransR\cite{lin2015learning}}
\chronoevent[markdepth=-130pt,date=false,mark=false]{2015}{$\bullet$ TransD\cite{ji2015knowledge}}
\chronoevent[markdepth=-120pt,date=false,mark=false]{2015}{$\bullet$ RTransE\cite{garcia2015composing}}
\chronoevent[markdepth=-110pt,date=false,mark=false]{2015}{$\bullet$ PTransE\cite{lin2015modeling}}
\chronoevent[markdepth=-100pt,date=false]{2015}{$\bullet$ KG2E\cite{he2015learning}}

\chronoevent[markdepth=-140pt,mark=false]{2016}{$\bullet$ FTransE\cite{feng2016knowledge}}
\chronoevent[markdepth=-130pt,date=false,mark=false]{2016}{$\bullet$ STransE\cite{nguyen2016stranse}}
\chronoevent[markdepth=-120pt,date=false,mark=false]{2016}{$\bullet$ TransG\cite{xiao2015transg}}
\chronoevent[markdepth=-110pt,date=false,mark=false]{2016}{$\bullet$ TransA\cite{jia2016locally}}
\chronoevent[markdepth=-100pt,date=false,mark=false]{2016}{$\bullet$ lppTransE\cite{yoon2016translation}}
\chronoevent[markdepth=-90pt,date=false,mark=false]{2016}{$\bullet$ TranSparse\cite{ji2016knowledge}}
\chronoevent[markdepth=-80pt,date=false]{2016}{$\bullet$ ManifoldE\cite{xiao2015one}}

\chronoevent[markdepth=-140pt,mark=false]{2017}{$\bullet$ puTransE\cite{tay2017non}}
\chronoevent[markdepth=-130pt,date=false,mark=false]{2017}{$\bullet$ ITransF\cite{xie2017interpretable}}
\chronoevent[markdepth=-120pt,date=false,mark=false]{2017}{$\bullet$ CombinE\cite{tan2017representation}}
\chronoevent[markdepth=-110pt,date=false]{2017}{$\bullet$ TransE-RS\cite{zhou2017learning}}

\chronoevent[markdepth=-140pt,mark=false]{2018}{$\bullet$ TransC\cite{lv2018differentiating}}
\chronoevent[markdepth=-130pt,date=false,mark=false]{2018}{$\bullet$ TransAt\cite{qian2018translating}}
\chronoevent[markdepth=-120pt,date=false]{2018}{$\bullet$ TorusE\cite{ebisu2018toruse}}

\chronoevent[markdepth=-140pt,mark=false]{2019}{$\bullet$ MuRP\cite{balazevic2019multi}}
\chronoevent[markdepth=-130pt,date=false,mark=false]{2019}{$\bullet$ QuatE\cite{zhang2019quaternion}}
\chronoevent[markdepth=-120pt,date=false,mark=false]{2019}{$\bullet$ RotatE\cite{sun2018rotate}}
\chronoevent[markdepth=-110pt,date=false,mark=false]{2019}{$\bullet$ KEC\cite{guan2019knowledge}}
\chronoevent[markdepth=-100pt,date=false,mark=false]{2019}{$\bullet$ TransGate\cite{yuan2019transgate}}
\chronoevent[markdepth=-90pt,date=false]{2019}{$\bullet$ TransMS\cite{yang2019transms}}

\chronoevent[markdepth=-140pt]{2020}{$\bullet$ ATTH\cite{chami2020low}}
\chronoevent[markdepth=-130pt,date=false,mark=false]{2020}{$\bullet$ HyperKG\cite{kolyvakis2020hyperbolic}}
\chronoevent[markdepth=-120pt,date=false,mark=false]{2020}{$\bullet$ Rotate3D\cite{gao2020rotate3d}}
\chronoevent[markdepth=-110pt,date=false,mark=false]{2020}{$\bullet$ GeomE\cite{xu2020knowledge}}
\chronoevent[markdepth=-100pt,date=false,mark=false]{2020}{$\bullet$ HAKE\cite{zhang2020learning}}
\chronoevent[markdepth=-90pt,date=false,mark=false]{2020}{$\bullet$ BoxE\cite{abboud2020boxe}}
\chronoevent[markdepth=-80pt,date=false,mark=false]{2020}{$\bullet$ OTE\cite{tang-etal-2020-orthogonal}}
\chronoevent[markdepth=-70pt,date=false,mark=false]{2020}{$\bullet$ RatE\cite{huang2020rate}}
\chronoevent[markdepth=-60pt,date=false,mark=false]{2020}{$\bullet$ MDE\cite{sadeghi2019mde}}
\chronoevent[markdepth=-50pt,date=false,mark=false]{2020}{$\bullet$ AprilE\cite{liu2020aprile}}
\chronoevent[markdepth=-40pt,date=false,mark=false]{2020}{$\bullet$ TransRHS\cite{zhang2020transrhs}}
\chronoevent[markdepth=-30pt,date=false,mark=false]{2020}{$\bullet$ SpacE\cite{nayyeri2020fantastic}}
\chronoevent[markdepth=-20pt,date=false,mark=false]{2020}{$\bullet$ LineaRE\cite{peng2020lineare}}

\chronoevent[markdepth=-140pt]{2021}{$\bullet$HBE\cite{pan2021hyperbolic}}
\chronoevent[markdepth=-130pt,date=false,mark=false]{2021}{$\bullet$RotL\cite{wang2021hyperbolic}}
\chronoevent[markdepth=-120pt,date=false,mark=false]{2021}{$\bullet$GrpKG\cite{yang2021knowledge}}
\chronoevent[markdepth=-110pt,date=false,mark=false]{2021}{$\bullet$HopfE\cite{bastos2021hopfe}}
\chronoevent[markdepth=-100pt,date=false,mark=false]{2021}{$\bullet$PairRE\cite{chao-etal-2021-pairre}}
\chronoevent[markdepth=-90pt,date=false,mark=false]{2021}{$\bullet$MRotatE\cite{huang2021knowledge}}
\chronoevent[markdepth=-80pt,date=false,mark=false,mark=false]{2021}{$\bullet$HA-RotatE\cite{wang2021hierarchical}}
\chronoevent[markdepth=-70pt,date=false,mark=false,mark=false,mark=false]{2021}{$\bullet$MQuadE\cite{yu2021mquade}}
\chronoevent[markdepth=-60pt,date=false,mark=false,mark=false,mark=false,mark=false]{2021}{$\bullet$CyclE\cite{yang2021cycle}}
\chronoevent[markdepth=-50pt,date=false,mark=false]{2021}{$\bullet$5E\cite{chen2021mobiuse}}
\chronoevent[markdepth=-40pt,date=false,mark=false]{2021}{$\bullet$DualE\cite{cao2021dual}}
\chronoevent[markdepth=-30pt,date=false,mark=false]{2021}{$\bullet$BiQUE\cite{guo2021bique}}
\chronoevent[markdepth=-20pt,date=false,mark=false]{2021}{$\bullet$FieldE\cite{nayyeri2021knowledge}}

\chronoevent[markdepth=-140pt]{2022}{$\bullet$ StructurE\cite{zhang2022structural}}
\chronoevent[markdepth=-130pt,date=false,mark=false]{2022}{$\bullet$ ReflectE\cite{zhang2022knowledge}}
\chronoevent[markdepth=-120pt,date=false,mark=false]{2022}{$\bullet$ DensE\cite{lu2022dense}}
\chronograduation{1}
\stopchronology
\caption{Timeline of Knowledge Graph Embedding models.}
\label{fig:kge_chrono}
\end{figure}

Knowledge Graphs (KGs) serve as vital repositories of information for many real-world applications and services, including search engines, virtual assistants, knowledge discovery, and fraud detection. The construction of KGs primarily involves domain expert curation or the automated extraction of data from vast web corpora. Despite the precision achieved by machine learning models in entity and relation extraction, they can introduce errors during KG construction. Furthermore, due to the inherent incompleteness of entity information, KG embedding (KGE) techniques come into play to identify missing relationships between entities. Over the past decade, there has been a surge in interest with the creation of various KGE models as evidenced in Figure \ref{fig:kge_chrono}. As such, it will be valuable to have an overview of extant KGE models to compare their similarities and differences, as well as a summary of research resources such as public KGs, benchmarking datasets, and leaderboards. In this paper, we will give a comprehensive overview of previous developments in KGE models. In particular, we will focus on distance-based and semantic matching KGE models. In recent development of KGE models, we have observed an interesting trend of combining different geometric transformations to improve the performance of existing KGE models. Basic transformations, including translation, rotation, scaling, reflection, and shear, are simple yet very powerful tools for representing relations between entities in KG. In this paper, we will also present how these tools can be combined to come up with more powerful models.

\subsection{Background}

\begin{figure}[ht!]
\centering
\includegraphics[width=0.35\columnwidth]{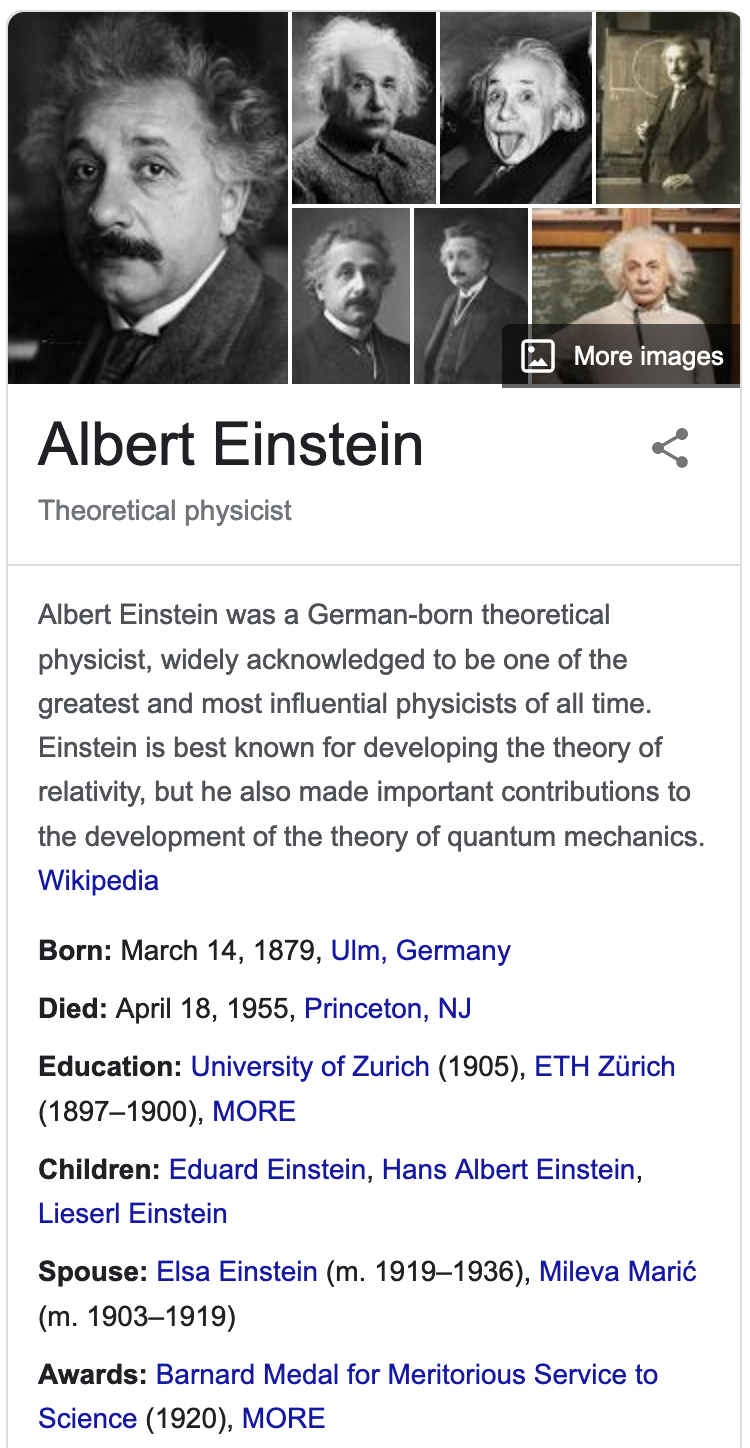}
\caption{Illustration of Knowledge Panel from Google Search.}\label{fig:knowledge_panel}
\end{figure}

KG finds vast and diverse applications. It enables swift retrieval of structured data about target entities during user searches. For instance, when you search for a well-known person, place, or popular topic on Google, the Google Knowledge Panel, shown in Figure \ref{fig:knowledge_panel}, accompanies search results, providing quick insights into the subject of interest. The data source for the Knowledge Panel is the Google KG, launched in 2012, initially derived from Freebase, an open-source KG acquired by Google in 2010, and later augmented with data from sources like Wikidata.

KG information integration extends to various domains, with E-commerce companies creating user and product KGs and merging them with other KGs to gain business intelligence. Hospitals and clinics employ KGs to share patient medical conditions, especially for patients relocating to different areas. Financial institutions use KGs to track illegal activities such as money laundering. Moreover, KGs serve as essential information sources for AI-powered virtual assistants like Siri, Alexa, and Google Assistant. Natural Language Understanding (NLU) algorithms analyze dialogs, extracting keywords to locate relevant KG subgraphs. By traversing these subgraphs, the assistants generate sensible responses using Natural Language Generation models. KGs also find applications in music recommendation systems, event forecasting based on temporal KGs, and more.

KG representation learning has been a subject of intensive research in recent years and remains a foundational challenge in Artificial Intelligence (AI) and Data Engineering. KGs are composed of triples, denoted as $(h,r,t)$, where $h$ and $t$ denote head and tail entities, while $r$ signifies the connecting relation. For instance, the statement "Los Angeles is located in the USA" is encapsulated as the triple (Los Angeles, \textbf{isLocatedIn}, USA). KGE plays a critical role in a range of downstream applications, including multihop reasoning \cite{shen2020modeling, du2021cogkr}, KG alignment \cite{chen2016multilingual, largeEA, ge2023typeea}, entity classification \cite{ge2022core, wang2023asyncet}.

The evaluation of KGE models often revolves around the link prediction task, assessing their ability to predict $t$ given $h$ and $r$, or $h$ given $r$ and $t$. The effectiveness of KGE models is determined by how closely their predictions align with the ground truth. Designing effective KGE models presents several challenges. First, KGE needs to be scalable since real-world KGs often contain millions of entities and relations. Designing embeddings that scale efficiently to handle large graphs is a significant challenge. Second, KGs are typically incomplete and subject to continuous updates. It is desirable for embedding models for handling missing data and capturing temporal dynamics and the history of relationships. Third, embeddings must also be expressive enough to capture the complexity of real-world relationships, such as 1-to-N, N-to-1, N-to-N, antisymmetric, transitive, and hierarchical relations and multi-modal data. Fourth, some entities and relations are rare. Embeddings should handle the long-tail distribution of data effectively. 

We have collected a list of survey papers as shown in Table \ref{tab:survey_papers}. Among these surveys, \cite{bengio2013representation, nickel2015review, wang2017knowledge, cai2018comprehensive, chen2020review, rossi2021knowledge, ji2021survey, cao2022knowledge} focus on discussing different embedding models, whereas \cite{hogan2021knowledge, ji2021survey, tiddi2022knowledge} discuss the use of KG for reasoning and different applications. \cite{gutierrez2021knowledge} elucidates the evolution of KGs and the reasons for their invention in a historical context. \cite{hogan2021knowledge} summarizes methods for the creation, enrichment, quality assessment, refinement, and publication of KGs, and provides an overview of prominent open KGs and enterprise KGs. \cite{tiddi2022knowledge} also discusses the advantages and disadvantages of using KGs as background knowledge in the context of Explainable Machine Learning. However, none of these papers discuss the intrinsic connections between different distance-based embedding models that use geometric transformations. We believe that this paper will be helpful to the research community by providing a unique perspective on this topic.

\begin{longtable}{c|c|p{4cm}|p{5cm}|p{4cm}}
\caption{Survey Papers.} \label{tab:survey_papers}\\
      \hline
      \textbf{No.} & \textbf{Year} & \textbf{Title} & \textbf{Author} & \textbf{Venue}\\
      \hline
      \raisebox{-1.5ex}[0pt]{\footnotesize 1 } & \footnotesize 2013 & \footnotesize Representation Learning: A Review and New Perspectives \cite{bengio2013representation} & \footnotesize Yoshua Bengio, Aaron C. Courville, Pascal Vincent. & \footnotesize IEEE Transactions on Pattern Analysis and Machine Intelligence\\
      \hline
      \raisebox{-1.5ex}[0pt]{\footnotesize 2 } & \footnotesize 2016 & \footnotesize A Review of Relational Machine Learning for Knowledge Graphs \cite{nickel2015review}  & \footnotesize Maximilian Nickel, Kevin Murphy, Volker Tresp, Evgeniy Gabrilovich. & \footnotesize Proceedings of the IEEE\\
      \hline
      \raisebox{-1.5ex}[0pt]{\footnotesize 3 } & \footnotesize 2017 & \footnotesize Knowledge Graph Embedding: A Survey of Approaches and Applications \cite{wang2017knowledge} & \footnotesize Quan Wang, Zhendong Mao, Bin Wang, Li Guo. & \footnotesize IEEE Transactions on Knowledge and Data Engineering\\
      \hline
      \raisebox{-1.5ex}[0pt]{\footnotesize 4 } & \footnotesize 2018 & \footnotesize A Comprehensive Survey of Graph Embedding: Problems, Techniques, and Applications. \cite{cai2018comprehensive}  & \footnotesize HongYun Cai, Vincent W. Zheng, Kevin Chen-Chuan Chang. & \footnotesize IEEE Transactions on Knowledge and Data Engineering\\
      \hline     
      \raisebox{-1.5ex}[0pt]{\footnotesize 5 } & \footnotesize 2020 & \footnotesize A review: Knowledge Reasoning over Knowledge Graph. \cite{chen2020review}  & \footnotesize Xiaojun Chen, Shengbin Jia, Yang Xiang. & \footnotesize Expert Systems with Applications\\
      \hline
      \raisebox{-0.5ex}[0pt]{\footnotesize 6 } & \footnotesize 2021 & \footnotesize Knowledge Graphs \cite{gutierrez2021knowledge} & \footnotesize Claudio Gutierrez, Juan F. Sequeda. & \footnotesize Communications of the ACM \\
      \hline
      \raisebox{-1.5ex}[0pt]{\footnotesize 7 } & \footnotesize 2021 & \footnotesize Knowledge Graph Embedding for Link Prediction: A Comparative Analysis \cite{rossi2021knowledge} & \footnotesize Rossi, Andrea and Barbosa, Denilson and Firmani, Donatella and Matinata, Antonio and Merialdo, Paolo. & \footnotesize ACM Transactions on Knowledge Discovery from Data  \\
      \hline
      \raisebox{-1.5ex}[0pt]{\footnotesize 8 } & \footnotesize 2021 & \footnotesize Knowledge Graphs \cite{hogan2021knowledge}  & \footnotesize Aidan Hogan, Eva Blomqvist, Michael Cochez, Claudia D’amato, Gerard De Melo, Claudio Gutierrez, Sabrina Kirrane, Jos\'e Emilio Labra Gayo, Roberto Navigli, Sebastian Neumaier, Axel-Cyrille Ngonga Ngomo, Axel Polleres, Sabbir M. Rashid,Anisa Rula, Lukas Schmelzeisen, Juan Sequeda, Steffen Staab, Antoine Zimmermann. & \footnotesize ACM Computing Surveys\\
      \hline
      \raisebox{-1.5ex}[0pt]{\footnotesize 9 } & \footnotesize 2022 & \footnotesize Knowledge Graphs: A Practical Review of the Research Landscape \cite{kejriwal2022knowledge} & \footnotesize Mayank Kejriwal. & \footnotesize Information \\
      \hline
      \raisebox{-1.5ex}[0pt]{\footnotesize 10 } & \footnotesize 2022 & \footnotesize Knowledge Graphs as Tools for Explainable Machine Learning: A survey \cite{tiddi2022knowledge} & \footnotesize Ilaria Tiddi, Stefan Schlobach. & \footnotesize Artificial Intelligence  \\
      \hline
      \raisebox{-1.5ex}[0pt]{\footnotesize 11 } & \footnotesize 2022 & \footnotesize A Survey on Knowledge Graphs: Representation, Acquisition, and Applications \cite{ji2021survey} & \footnotesize Shaoxiong Ji, Shirui Pan, Erik Cambria, Pekka Marttinen, Philip S. Yu. & \footnotesize IEEE Transactions on Neural Networks and Learning Systems \\
      \hline
      \raisebox{-1.5ex}[0pt]{\footnotesize 12 } & \footnotesize 2022 & \footnotesize Knowledge Graph Embedding: A Survey from the Perspective of Representation Spaces \cite{cao2022knowledge} & \footnotesize Jiahang Cao, Jinyuan Fang, Zaiqiao Meng, Shangsong Liang. & \footnotesize arXiv prepints \\
      \hline 
\end{longtable}


\subsection{Our Contributions}

Our main contributions in this paper can be summarized as follows.

\begin{itemize}
\item We review different KGE models, focusing on distance-based and semantic matching models.
\item We collect relevant resources for KG research, including previously published survey papers, major open 
sourced KGs, and KG benchmarking datasets for link prediction, as well as link prediction performance on some of the most popular datasets.
\item We discover the connections between recently published KGE models and propose CompoundE and CompoundE3D, which follow this direction of thought.
\item We discuss recent work that leverages neural network models such as graph neural networks and pretrained language models, and how embedding-based models can be combined with neural network-based models.

\end{itemize}

The rest of this paper is organized as follows. In Section \ref{sec:existing_models}, we first introduce existing KGE models in both distance-based and semantic matching-based categories. We also discuss a number of commonly used loss functions and their suitability for different types of KGE scoring functions. In Section \ref{sec:unified_framwork}, we present CompoundE, followed by CompoundE3D, which unifies all distance-based KGE models that use affine operations. In Section \ref{sec:datasets_and_evaluation}, we summarize a list of open sourced KGs, popular benchmarking datasets for performance evaluation, and the evaluation metrics used in these datasets. We also provide the recent leaderboard for some popular datasets. In Section \ref{sec:emerging}, we discuss existing neural network-based models for KG completion and emerging directions that use pretrained language models. Finally, in Section \ref{sec:conclusion}, we make concluding remarks.


\begin{table}
    \caption{Distance-based KGE models}
    \label{tab:translation_kge}
    \centering
    \begin{adjustbox}{width=\textwidth,center}
    \begin{tabular}{ccccc}
        \hline
        Model & Ent. emb. & Rel. emb. & Scoring Function & Space \\
        \hline
        TransE~\cite{bordes2013translating} & $\mathbf{h}, \mathbf{t}\in\mathbb{R}^d$ & $\mathbf{r}\in\mathbb{R}^d$ & $-\left\|\mathbf{h}+\mathbf{r}-\mathbf{t}\right\|_{1/2}$ & $O(md+nd)$ \\
        TransR~\cite{lin2015learning}	&	$\mathbf{h}, \mathbf{t} \in \mathbb{R}^{d}$	&	$\mathbf{r} \in \mathbb{R}^{k}, \mathbf{M}_{r} \in \mathbb{R}^{k \times d}$	&	$-\left\|\mathbf{M}_{r} \mathbf{h}+\mathbf{r}-\mathbf{M}_{r} \mathbf{t}\right\|_{2}^{2}$ & $O(mdk+nd)$\\
        TransH~\cite{wang2014knowledge}	&	$\mathbf{h}, \mathbf{t} \in \mathbb{R}^{d}$	&	$\mathbf{r}, \mathbf{w}_{r} \in \mathbb{R}^{d}$	&	$-\left\|\left(\mathbf{h}-\mathbf{w}_{r}^{\top} \mathbf{h} \mathbf{w}_{r}\right)+\mathbf{r}-\left(\mathbf{t}-\mathbf{w}_{r}^{\top} \mathbf{t} \mathbf{w}_{r}\right)\right\|_{2}^{2}$ & $O(md+nd)$	\\
	    TransA~\cite{jia2016locally}	&	$\mathbf{h}, \mathbf{t} \in \mathbb{R}^{d}$	&	$\mathbf{r} \in \mathbb{R}^{d}, \mathbf{M}_{r} \in \mathbb{R}^{d \times d}$	&	$(|\mathbf{h}+\mathbf{r}-\mathbf{t}|)^{\top} \mathbf{W}_{\mathbf{r}}(|\mathbf{h}+\mathbf{r}-\mathbf{t}|)$ & $O(md^2+nd)$ \\
        TransF~\cite{feng2016knowledge} 	&	$\mathbf{h}, \mathbf{t} \in \mathbb{R}^{d}$	&	$\mathbf{r} \in \mathbb{R}^{d}$	&	$(\mathbf{h}+\mathbf{r})^{\top} \mathbf{t}+(\mathbf{t}-\mathbf{r})^{\top} \mathbf{h}$ & $O(md+nd)$ \\
	    TransD~\cite{ji2015knowledge}	&	$\mathbf{h}, \mathbf{h}_{p}, \mathbf{t}, \mathbf{t}_{p} \in \mathbb{R}^{d}$	&	$\mathbf{r}, \mathbf{r}_{p} \in \mathbb{R}^{k}$	&	
	    $-\left\|\left(\mathbf{r}_{p} \mathbf{h}_{p}^{T}+\mathbf{I}\right) \mathbf{h}+\mathbf{r}-\left(\mathbf{r}_{p} \mathbf{t}_{p}^{T}+\mathbf{I}\right) \mathbf{t}\right\|_{2}^{2}$ & $O(mk+nd)$	\\
	    TransM~\cite{fan2014transition}	&	$\mathbf{h}, \mathbf{t} \in \mathbb{R}^{d}$	&	$\mathbf{r} \in \mathbb{R}^{d}$, $w_{r}\in\mathbb{R}$&	$-w_{r}\|\mathbf{h}+\mathbf{r}-\mathbf{t}\|_{1 / 2}$  & $O(md+nd)$	\\
	    \multirow{2}{70pt}{TranSparse~\cite{ji2016knowledge}}	&	\multirow{2}{50pt}{$\mathbf{h}, \mathbf{t} \in \mathbb{R}^{d}$}	&	$\mathbf{r} \in \mathbb{R}^{k}, \mathbf{M}_{r}\left(\theta_{r}\right) \in \mathbb{R}^{k \times d}$	&	$-\left\|\mathbf{M}_{r}\left(\theta_{r}\right) \mathbf{h}+\mathbf{r}-\mathbf{M}_{r}\left(\theta_{r}\right) \mathbf{t}\right\|_{1 / 2}^{2}$  & \multirow{2}{70pt}{$O(mdk+nd)$} \\
	    &		&	$\mathbf{M}_{r}^{1}\left(\theta_{r}^{1}\right), \mathbf{M}_{r}^{2}\left(\theta_{r}^{2}\right) \in \mathbb{R}^{k \times d}$	&	$-\left\|\mathbf{M}_{r}^{1}\left(\theta_{r}^{1}\right) \mathbf{h}+\mathbf{r}-\mathbf{M}_{r}^{2}\left(\theta_{r}^{2}\right) \mathbf{t}\right\|_{1 / 2}^{2}$	\\
        ManifoldE~\cite{xiao2015one}	&	$\mathbf{h}, \mathbf{t} \in \mathbb{R}^{d}$	&	$\mathbf{r} \in \mathbb{R}^{d}$	&	$\left\|\mathcal{M}(h, r, t)-D_{r}^{2}\right\|^{2}$	& $O(md+nd)$\\
	    TorusE~\cite{ebisu2018toruse}	&	$[\mathbf{h}], [\mathbf{t}] \in \mathbb{T}^{n}$	&	$[\mathbf{r}] \in \mathbb{T}^{n}$	&	 $\min _{(x, y) \in([h]+[r]) \times[t]}\|x-y\|_{i}$ & $O(md+nd)$\\
        RotatE~\cite{sun2018rotate} & $\mathbf{h}, \mathbf{t}\in\mathbb{C}^d$ & $\mathbf{r}\in\mathbb{C}^d$ & $-\left\|\mathbf{ h \circ r - t }\right\|$ & $O(md+nd)$ \\
        PairRE~\cite{chao-etal-2021-pairre} & $\mathbf{h}, \mathbf{t}\in\mathbb{R}^d$ & $\mathbf{r^H, r^T}\in\mathbb{R}^d$ & $-\|\mathbf{h\circ r}^H - \mathbf{t\circ r}^T\|$ & $O(md+nd)$ \\
        \hline
    \end{tabular}
    \end{adjustbox}
\end{table}

\begin{table}
    \caption{Semantic matching-based KGE models. }
    \label{tab:geometry_kge}
    \centering
    \begin{adjustbox}{width=0.9\textwidth,center}
    \begin{tabular}{ccccc}
        \hline
        Model & Ent. emb. & Rel. emb. & Scoring Function & Space \\
        \hline
        RESCAL \cite{nickel2011three} &	$\mathbf{h}, \mathbf{t} \in \mathbb{R}^{d}$	&	$\mathbf{M}_{r} \in \mathbb{R}^{d \times d}$	&	$\mathbf{h}^{\top}\mathbf{M}_r\mathbf{t}$ &	$O(md^2+nd)$\\
        DistMult \cite{yang2014embedding} &	$\mathbf{h}, \mathbf{t} \in \mathbb{R}^{d}$	&	$\mathbf{r} \in \mathbb{R}^{d}$	&	$\mathbf{h}^{\top} \operatorname{diag}(\mathbf{r}) \mathbf{t}$ & $O(md+nd)$\\
        HolE \cite{nickel2016holographic} & $\mathbf{h}, \mathbf{t} \in \mathbb{R}^{d}$	&	$\mathbf{r} \in \mathbb{R}^{d}$	&	$\mathbf{r}^\top (h\star t)$ & $O(md+nd)$\\
        ANALOGY \cite{liu2017analogical} &	$\mathbf{h}, \mathbf{t} \in \mathbb{R}^{d}$	&	$\mathbf{\hat{M}}_r \in \mathbb{R}^{d \times d}$	&	$ \mathbf{h}^{\top} \mathbf{\hat{M}}_{r} \mathbf{t}$ & $O(md+nd)$\\
        ComplEx \cite{trouillon2016complex} & $\mathbf{h,t}\in\mathbb{C}^d$ & $\mathbf{r}\in\mathbb{C}^d$ & $\operatorname{Re}\left(\langle \mathbf{r}, \mathbf{h}, \mathbf{\overline{t}} \rangle\right)$ & $O(md+nd)$\\
        SimplE \cite{kazemi2018simple} &	$\mathbf{h}, \mathbf{t} \in \mathbb{R}^d$	&	$\mathbf{r}, \mathbf{r}^\prime\in\mathbb{R}^d$	&	 $\frac{1}{2}\left(\langle \mathbf{h},\mathbf{r}, \mathbf{t}\rangle + \langle \mathbf{t}, \mathbf{r}^\prime, \mathbf{h}\rangle\right)$ & $O(md+nd)$\\
        Dihedral \cite{xu2019relation} &	$\mathbf{h}^{(l)}, \mathbf{t}^{(l)} \in \mathbb{R}^{2}$	&	$\mathbf{R}^{(l)} \in \mathbb{D}_{K}$	&	$\sum_{l=1}^{L} \mathbf{h}^{(l) \top} \mathbf{R}^{(l)} \mathbf{t}^{(l)}$ & $O(md+nd)$\\
        TuckER \cite{balavzevic2019tucker} &	$\mathbf{h}, \mathbf{t} \in \mathbb{R}^d_e$	&	$\mathbf{r}\in\mathbb{R}^d_r$	&	$\mathcal{W} \times_{1} \mathbf{h} \times_{2} \mathbf{r} \times_{3} \mathbf{t}$& $O(d_r d_e^2 + md_r+nd_e)$\\
        QuatE \cite{zhang2019quaternion} & $Q_h, Q_t \in \mathbb{H}^d$ & $W_r \in \mathbb{H}^d$ & $Q_h \otimes W_r^{\triangleleft} \cdot Q_t$ & $O(md+nd)$\\
        DualE \cite{cao2021dual} & $Q_h, Q_t \in \mathbb{H}^d$ & $W_r, T_r \in \mathbb{H}^d$ & $(Q_h \otimes W_r^{\triangleleft} + T_r)\cdot Q_t$ & $O(md+nd)$\\
        CrossE \cite{zhang2019interaction} & $\mathbf{h}, \mathbf{t} \in \mathbb{R}^d$	&	$\mathbf{r}\in\mathbb{R}^d$	&	$\sigma\left(\tanh \left(\mathbf{c}_{r} \circ \mathbf{h}+\mathbf{c}_{r} \circ \mathbf{h} \circ \mathbf{r}+\mathbf{b}\right) \mathbf{t}^{\top}\right)$ & $O(md+nd)$\\
        SEEK \cite{xu2020seek} & $\mathbf{h}, \mathbf{t} \in \mathbb{R}^d$	&	$\mathbf{r}\in\mathbb{R}^d$ & $\sum s_{x, y} \left\langle \mathbf{r}_x, \mathbf{h}_y, \mathbf{t}_{w_{x, y}} \right\rangle$ & $O(md+nd)$\\
        \hline
    \end{tabular}
    \end{adjustbox}
\end{table}

\section{Existing Models}\label{sec:existing_models}

KGE models are often categorized based on scoring functions and tools applied to model entity-relation interactions and representations. In this paper, we mainly discuss two major classes, namely 1) distance-based models, and 2) semantic matching models. 

\subsection{Distance-based Models}

Distance-based scoring function is one of the most popular strategies for learning KGE. The intuition behind this strategy is that relations are modeled as transformations to place head entity vectors in the proximity of their corresponding tail entity vectors or vice versa. For a given triple $(h,r,t)$, the goal is to minimize the distance between $h$ and $t$ vectors after the transformation introduced by $r$. 

TransE \cite{bordes2013translating} is one of the first KGE models that interpret relations between entities as translation operations in vector space. Let $\mathbf{h,r,t}\in\mathbb{R}^d$ denote the embedding for head, relation, and tail of a triple, respectively. TransE scoring function is defined as:
\begin{equation}
    f_r(h,r) = \|\mathbf{h+r-t}\|_{p},
\end{equation}
where $p=1$ or $2$ denote 1-Norm or 2-Norm, respectively. However, this efficient model has difficulty modeling complex relations such as 1-N, N-1, N-N, symmetric and transitive relations. Many later works attempt to overcome this shortcoming. For example, TransH \cite{wang2014knowledge} projects entity embedding onto relation-specific hyperplanes so that complex relations can be modeled by the translation embedding model. Formally, let $\mathbf{w}_r$ be the normal vector to a relation-specific hyperplane, then the head and tail representation in the hyperplane can be written as,
\begin{equation}
    \mathbf{h}_\bot = \mathbf{h}-\mathbf{w}_r^{\top}\mathbf{h}\mathbf{w}_r,\quad\mathbf{t}_\bot = \mathbf{t}-\mathbf{w}_r^{\top}\mathbf{t}\mathbf{w}_r.
\end{equation}
The projected representations are then linked together using the same translation relationship,
\begin{equation}
    f_r(h,r) =\|\mathbf{h}_\bot+\mathbf{r}-\mathbf{t}_\bot\|_2^2.
\end{equation}
However, this orthogonal projection prevents the model from encoding inverse and composition relations. A similar idea called TransR \cite{lin2015learning} transforms entities into a relation-specific space instead. The TransR scoring function can be written as,
\begin{equation}
    f_r(h,r) =\|\mathbf{M}_r\mathbf{h}+\mathbf{r}-\mathbf{M}_r\mathbf{t}\|_2^2.    
\end{equation}
However, the relation-specific transformation introduced in TransR requires $O(kd)$ additional parameters. 
To save the additional parameters introduced, TransD \cite{ji2015knowledge} uses entity projection vectors to populate the mapping matrices, instead of using a dense matrix. TransD reduces the additional parameters from $O(kd)$ to $O(k)$. The scoring function can be written as,
\begin{equation}
    f_r(h,r) =\left\|\left(\mathbf{r}_{p} \mathbf{h}_{p}^{\top}+\mathbf{I}\right) \mathbf{h}+\mathbf{r}-\left(\mathbf{r}_{p} \mathbf{t}_{p}^{\top}+\mathbf{I}\right) \mathbf{t}\right\|_{2}^{2}.
\end{equation}
With the same goal of saving additional parameters, TranSparse enforces the transformation matrix to be a sparse matrix. The scoring function can be written as,

\begin{equation}
    f_r(h,t) = \left\|\mathbf{M}_{r}\left(\theta_{r}\right) \mathbf{h}+\mathbf{r}-\mathbf{M}_{r}\left(\theta_{r}\right) \mathbf{t}\right\|_{1 / 2}^{2},
\end{equation}
where $\theta_r\in [0,1]$ is the sparse degree for the mapping matrix $\mathbf{M}_r$. Variants of TranSparse \cite{ji2016knowledge} include separate mapping matrices for head and tail. TransM \cite{fan2014transition} assigns different weights to complex relations for better encoding power. TransMS \cite{yang2019transms} attempts to consider multidirectional semantics using nonlinear functions and linear bias vectors. TransF \cite{feng2016knowledge} mitigates the burden of relation projection by explicitly modeling the basis of projection matrices. ITransF \cite{xie2017interpretable} makes use of concept projection matrices and sparse attention vectors to discover hidden concepts within relations.

In recent years, researchers expand their focus to spaces other than Euclidean geometry. TorusE \cite{ebisu2019generalized} projects embedding in an n-dimensional torus space, where $[\mathbf{h}], [\mathbf{r}], [\mathbf{t}] \in \mathbb{T}^{n}$ denotes the projected representation of head, relation, tail. TorusE models relational translation in Torus space by optimizing the objective as follows. 
\begin{equation}
    \min _{(x, y) \in([h]+[r]) \times[t]}\|x-y\|_{i}.
\end{equation}

Multi-Relational Poincar\'e model (MuRP) \cite{balazevic2019multi} embeds KG entities in a Poincar\'e ball of hyperbolic space. It transforms entity embeddings using relation-specific M\"obius matrix-vector multiplication and M\"obius addition. The negative curvature introduced by hyperbolic space is empirically better in capturing the hierarchical structure in KGs. However, MuRP has difficulty encoding relation patterns and only uses a constant curvature. ROTH \cite{chami2020low} improve over MuRP by introducing a relation-specific curvature. 

RotatE \cite{sun2018rotate} models entities in the complex vector space and interprets relations as rotations instead of translations. Formally, let $\mathbf{h}, \mathbf{r}, \mathbf{t}\in\mathbb{C}^d$ denote the representation of head, relation, and tail of a triple in the complex vector space. The RotatE scoring function can be defined as,

\begin{equation}
    f_r(h,t) = \left\|\mathbf{ h \circ r - t }\right\|.
\end{equation}
The self-adversarial negative sampling strategy also contributes to RotatE's significant performance improvement compared to its predecessors. Quite a few models attempt to extend RotatE. MRotatE adds an entity rotation constraint to the optimization objective to handle multifold relations. HAKE rewrites the rotation formula in polar coordinates and separates the scoring function into two components, that is, the phase component and the modulus component. The scoring function of HAKE can be written as,

\begin{equation}
    f_r(h,t) = d_{r,m}(\mathbf{h,t}) + \lambda d_{r,p}(\mathbf{h,t}),
\end{equation}

where 
\begin{equation}
    d_{r,p}(\mathbf{h,t}) = \|\sin((\mathbf{h}_p+\mathbf{r}_p-\mathbf{t}_p)/2)\|_1,
\end{equation}
and
\begin{equation}
    d_{r,m}(\mathbf{h,t}) = \|\mathbf{h}_m\circ((\mathbf{r}_m+\mathbf{r}_m^\prime)/(1-\mathbf{r}_m^\prime))-\mathbf{t}_m\|_2.
\end{equation}
This modification leads to better modeling capability of hierarchy structures in KG. Rotate3D performs quaternion rotation in 3D space and enables the model to encode non-commutative relations. Rot-Pro extends the RotatE by transforming entity embeddings using an orthogonal projection that is also idempotent. This change enables RotPro to model transitive relations. PairRE also tries to improve over RotatE. Instead of rotating the head to match the tail, PairRE \cite{chao-etal-2021-pairre} performs transformations on both head and tail. The scoring function can be defined as,
\begin{equation}
   f_r(h,t) = \|\mathbf{h\circ r^H - t\circ r^T}\|,
\end{equation}
where $\mathbf{h,t}\in \mathbb{R}^d$ are head and tail entity embedding, and $\mathbf{r^H}, \mathbf{r^T}\in \mathbb{R}^d$ are relation-specific weight vectors for head and tail vectors respectively, and $\circ$ is an elementwise product. In fact, this elementwise multiplication is simply the scaling operation. One advantage of PairRE compared to previous models is that it is capable of modeling subrelation structures in KG. LinearRE \cite{peng2020lineare} is a similar model but adds a translation component between the scaled head and tail embedding. The transformation strategy can still be effective by adding it to entity embedding involved in relation rotation. SFBR \cite{liang2021semantic} introduces a semantic filter which includes a scaling and shift component. HousE \cite{li2022house} and ReflectE \cite{zhang2022knowledge} models relation as Householder reflection. UltraE \cite{xiong2022ultrahyperbolic} unifies Euclidean and hyperbolic geometry by modeling each relation as a pseudo-orthogonal transformation that preserves the Riemannian bilinear form. On the other hand, RotL \cite{wang2021hyperbolic} investigates the necessity of introducing hyperbolic geometry in learning KGE and proposes two more efficient Euclidean space KGE while retaining the advantage of flexible normalization.

\subsection{Semantic Matching Models}

Another related idea of developing KGE models is to measure the semantic matching score. RESCAL \cite{nickel2011three} adopts a bilinear scoring function as the objective in solving a three-way rank-$r$ matrix factorization problem. Formally, let $\mathbf{h,t}\in\mathbb{R}^d$ denote the head and tail embedding and $\mathbf{M}_r\in\mathbb{R}^{d\times d}$ is the representation for relation. Then, the RESCAL scoring function can be defined as,
\begin{equation}
    f_r(h,t) = \mathbf{h}^{\top}\mathbf{M}_r\mathbf{t}.
\end{equation}
However, one obvious limitation of this approach is that it uses a dense matrix to represent each relation, which requires an order of magnitude more parameters compared to those using vectors. DistMult \cite{yang2014embedding} reduces free parameters by enforcing the relation embedding matrix to be diagonal. Let $\mathbf{r}\in\mathbb{R}^d$ be the relation vector. Then, $\text{diag}(\mathbf{r})\in\mathbb{R}^{d\times d}$ is the diagonal matrix constructed from $\mathbf{r}$. Then, the DistMult scoring function can be written as,
\begin{equation}
    f_r(h,t) = \mathbf{h}^{\top}\text{diag}(\mathbf{r})\mathbf{t}.
\end{equation}
However, because the diagonal matrix is symmetric, it has difficulty modeling antisymmetric relations. ANALOGY \cite{liu2017analogical} has the same scoring function as RESCAL but instead it attempts to incorporate antisymmetric configurations by imposing two regularization constraints: 1) $\mathbf{M}_r\mathbf{M}_r^{\top} = \mathbf{M}_r^{\top}\mathbf{M}_r$ which requires the relation matrix to be orthonormal; 2) $\mathbf{M}_r\mathbf{M}_{r^\prime}=\mathbf{M}_{r^\prime}\mathbf{M}_r$ which requires the relation matrix to be commutative. HolE \cite{nickel2016holographic} introduces circular correlation between head and tail vectors, which can be interpreted as a compressed tensor product to capture richer interactions. The HolE scoring function can be written as,
\begin{equation}
    f_r(h,t) = \mathbf{r}^{\top}(\mathbf{h}\star\mathbf{t}).
\end{equation}
ComplEx \cite{trouillon2016complex} extends the bilinear product score to the complex vector space so as to model antisymmetric relations more effectively. Formally, let $\mathbf{h,r,t}\in\mathbb{C}^d$ be the head, relation, tail complex vectors, and $\overline{\mathbf{t}}$ denote the complex conjugate of the $\mathbf{t}$. The ComplEx scoring function can be defined as,
\begin{equation}
    f_r(h,t) = \operatorname{Re}(\langle \mathbf{r}, \mathbf{h}, 
\mathbf{\overline{t}} \rangle).
\end{equation}
where $\langle \cdot, \cdot, \cdot \rangle$ denotes trilinear product, and $\operatorname{Re}(\cdot)$ means taking the real part of a complex value. However, relation compositions remain difficult for ComplEx to encode. SimplE \cite{kazemi2018simple} models inverse of relations with an enhanced version of Canonical Polyadic decomposition. The scoring function of SimplE is defined as,
\begin{equation}
    f_r(h,t) = \frac{1}{2}\left(\langle \mathbf{h},\mathbf{r}, \mathbf{t}\rangle + \langle \mathbf{t}, \mathbf{r}^\prime, \mathbf{h}\rangle\right).
\end{equation}
TuckER \cite{balavzevic2019tucker} extends the semantic matching model to 3D tensor factorization of the binary tensor representation of KG triples. The scoring function is defined as,
\begin{equation}
    f_r(h,t) = \mathcal{W}\times_1 \mathbf{h} \times_2 \mathbf{r} \times_3 \mathbf{t}.
\end{equation}
QuatE \cite{zhang2019quaternion} and DualE \cite{cao2021dual} extend from the complex representation to the hypercomplex representation with 4 degrees of freedom to gain more expressive rotational capability. Let $Q_h, W_r, Q_t \in \mathbb{H}^k$ be the representation of head, relation, and tail in quaternion space of the form $Q=a+b\mathbf{i}+c\mathbf{j}+d\mathbf{d}$. Then the QuatE scoring function is defined as,
\begin{equation}
    f_r(h,t) = Q_h \otimes W_r^{\triangleleft} \cdot Q_t.
\end{equation}
Specifically, the normalization of relation vector in quaternion space is defined as,
\begin{equation}
    W_r^{\triangleleft}(p, q, u, v) = \frac{W_r}{|W_r|} = \frac{ a_r + b_r \textbf{i} + c_r \textbf{j} + d_r \textbf{k} }{\sqrt{a_r^2 + b_r^2 + c_r^2 + d_r^2}}.
\end{equation}
And the Hamiltonian product in quaternion space is computed as,
\begin{equation}
\label{hamilton}
\begin{split}
    Q_h \otimes W_r^{\triangleleft}   &=
   ( a_h \circ p - b_h\circ q - c_h\circ u - d_h\circ v )\\ &+ (a_h\circ q + b_h\circ p + c_h\circ v - d_h\circ u) \textbf{i} \\
   &+ (a_h\circ u - b_h\circ v + c_h\circ p + d_h\circ q) \textbf{j} \\ &+ (a_h\circ v + b_h\circ u - c_h\circ q + d_h\circ p) \textbf{k}.
\end{split}
\end{equation}
And the inner product in quaternion space is computed as,
\begin{equation}
Q_1 \cdot Q_2 = \langle a_1, a_2\rangle + \langle b_1,  b_2\rangle + \langle c_1, c_2\rangle + \langle d_1, d_2\rangle.
\end{equation}
However, one disadvantage of these models is that they require very high dimensional spaces to work well and therefore it is difficult to scale to large KGs. CrossE introduces crossover interactions to better represent the birdirectional interactions between entity and relations. The scoring function of CrossE is defined as,
\begin{equation}
   f_r(h,t) =  \sigma\left(\tanh \left(\mathbf{c}_{r} \circ \mathbf{h}+\mathbf{c}_{r} \circ \mathbf{h} \circ \mathbf{r}+\mathbf{b}\right) \mathbf{t}^{\top}\right),
\end{equation}
where the relation specific interaction vector $\mathbf{c}_r$ is obtained by looking up interaction matrix $\mathbf{C}\in\mathbb{R}^{n_r\times d}$. Diehdral \cite{xu2019relation} construct elements in a dihedral group using rotation and reflection operations over a 2D symmetric polygon. The advantage of the model is with encoding relation composition. SEEK \cite{xu2020seek} and AutoSF \cite{zhang2020autosf} identify the underlying similarities among popular KGEs and propose an automatic framework of designing new bilinear scoring functions while also unifying many previous models. However, the search space of AutoSF is computationally intractable and it is difficult to know if one configuration will be better than another unless the model is trained and tested with the dataset. Therefore, the AutoSF search can be time-consuming.

\subsection{Loss Functions}

Loss function is an important part of KGE learning. Loss functions are designed to effectively distinguish valid triples from negative samples. The ultimate goal of optimizing the loss function is to get valid triples ranked as high as possible. In early days of KGE learning, margin-based ranking loss is widely adopted. The pairwise max-margin loss can be formally defined as,

\begin{equation}
    L_R = \sum_{\substack{(h,r,t)\in \mathcal{G}\\(h',r,t')\in \mathcal{G'}}} \max(0, \gamma+f_r(h,t)-f_r(h',t')),
\end{equation}
where $(h,r,t)$ denotes ground truth triple from the set of all valid triples $\mathcal{G}$, $(h',r,t')$ denotes negative sample from the set of corrupted triples $\mathcal{G'}$. $\gamma$ is the margin parameter which specifies how different $f_r(h,t)$ and $f_r(h',t')$ should be at optimum. In fact, a similar loss function is applied to optimize multiclass Support Vector Machine (SVM) \cite{weston1999support}. Both distance-based embedding models, such as TransE, TransH, TransR, and TransD, and semantic matching-based models, such as LFM, NTN, and SME have successfully leveraged this scoring function. \cite{zhou2017learning} proposes a Limit-based scoring loss to limit the score of positive triples so that the translation relation in positive triples can be guaranteed. The Limit-based score can be defined as,

\begin{equation}
    L_{RS} = \sum_{\substack{(h,r,t)\in \mathcal{G}\\(h',r,t')\in \mathcal{G'}}} \{[\gamma+f_r(h,t)-f_r(h',t')]_+  +\lambda [f_r(h,t)-\mu]_+\}.
\end{equation}

More recently, a Double Limit Scoring Loss is proposed by \cite{zhou2021knowledge} to independently control the golden triplets' scores and negative samples' scores. It can be defined as,

\begin{equation}
    L_{SS} = \sum_{\substack{(h,r,t)\in \mathcal{G}\\(h',r,t')\in \mathcal{G'}}} \{[f_r(h,t)-\mu_p]_+ +\lambda [\mu_n - f_r(h',t')]_+\},
\end{equation}
where $\mu_n > \mu_p > 0$. This loss function intends to encourage low distance score for positive triplets and high distance scores for negative triplets. We can also trace the usage of a similar contrastive loss from Deep Triplet Network \cite{hoffer2015deep} for different image classification tasks.

Self adversarial negative sampling was proposed in RotatE \cite{sun2018rotate} and can be defined as,

\begin{equation}
     L_{SANS}=-\log\sigma(\gamma-f_r(h, t))-\sum_{i=1}^np(h'_i,r,t'_i)\log\sigma(f_r(h'_i,t'_i)-\gamma).
\end{equation}

Cross entropy or negative log-likelihood of logistic models are often used in semantic matching models where a product needs to be computed. The negative log-likelihood loss function can be defined as,

\begin{equation}
    L_{CE} = \sum_{(h,r,t)\in \mathcal{G}\cup\mathcal{G'}} \{1+\exp[-y_{(h,r,t)}\cdot f_r(h,t)]\}.
\end{equation}

Binary cross entropy or Bernoulli negative log-likelihood of logistic is also a popular loss function which can be defined as,

\begin{equation}
    L_{BCE} = \frac{1}{N_e} \sum_{i=1}^{N_e} y_i \log p_i + (1-y_i) \log (1-p_i).
\end{equation}

The binary cross entropy scoring function is more suitable for neural network-based models such as ConvE. Although TuckER is a semantic matching-based model, it also uses binary cross entropy as its loss function because its implementation is similar to a neural network.

\begin{figure}[t]
\centering
\includegraphics[width=\textwidth]{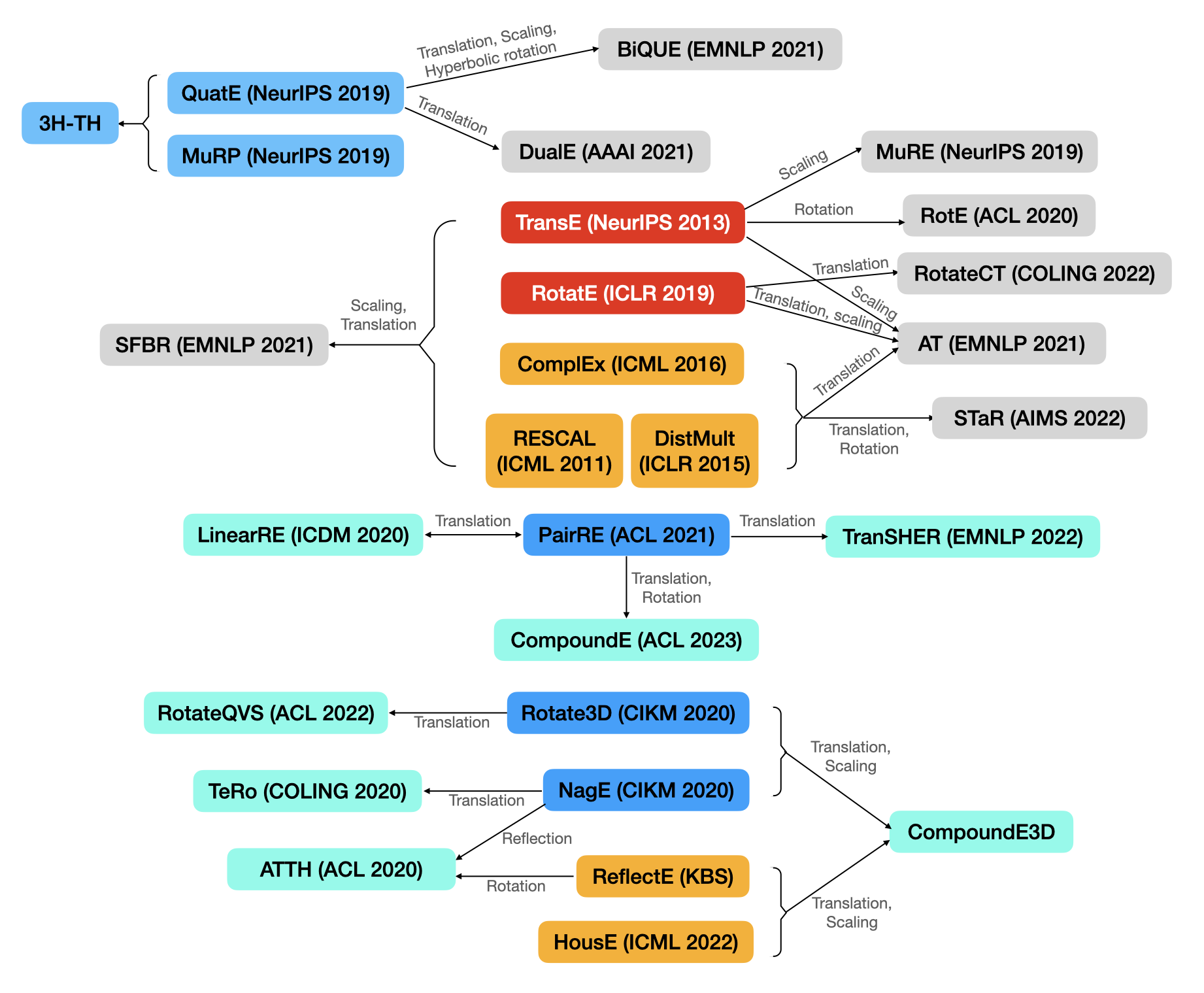}
\caption{Connections between different knowledge graph embedding models}
\label{fig:connections_between_kge}
\end{figure}

\subsection{Connections between Distance-based KGE Models}

Another interesting finding is that we discover the connections between different recent embedding models. In recent years, there is a trend to design more effective KGE using affine transformations. In fact, models are related to each other, and many of the new embedding models emerge from models that only use the most fundamental operations, such as translation (TransE), rotation (RotatE), and scaling (PairRE). We illustrate connections between recent KGE models in Figure \ref{fig:connections_between_kge}. For example, MuRE is a Euclidean version of MuRP, a popular hyperbolic space KGE model proposed in \cite{balazevic2019multi}. In MuRE, a diagonal matrix $\mathbf{R}\in \mathbb{R}^{d\times d}$ is applied to the head entity, and a translation vector $\mathbf{r}\in \mathbb{R}^{d}$ to the tail entity. The scoring function can be written as,
\begin{equation}
    \phi (e_s, r, e_o) = -d(\mathbf{Re_s,e_o+r})^2 + b_s+b_o,
\end{equation}
where $b_s$ and $b_o$ are the entity-specific bias terms for head and tail entities, respectively. This is essentially combining translation and scaling operations and MuRP implements a similar idea in the hyperbolic space. Similarly in \cite{chami2020low}, RotE and RotH are baseline models proposed in Euclidean space and hyperbolic space, respectively, that essentially apply 2D rotation operators to head entity embedding in the translational distance scoring function. The RotE scoring function can be defined as,
\begin{equation}
    s(h,r,t) = d(\text{Rot}(\mathbf{\Theta})\mathbf{e}_h+\mathbf{r}_r,\mathbf{e}_t) + b_h + b_t
\end{equation}
RefE and RefH can be derived similarly by applying 2D reflection operators. More recently, \cite{yang2021improving} combines translation and scaling operations with both distance-based models (TransE, RotatE) and semantic matching models (DistMult, ComplEx). Performance improvement in link prediction can be observed following this approach. SFBR \cite{liang2021semantic} applies semantic filters to distance-based and semantic matching-based models. One of the more effective MLP-based filters has diagonal weight matrices, which essentially apply scaling and translation operations to the entity embeddings. STaR \cite{li2022star} focuses on designing bilinear product matrices of semantic matching-based models by inserting scaling and translation components in order to enable the KGE to handle complex relation types such as 1-N, N-1, and N-N relations. 

Similar trend has also been observed in quaternion space KGE that was first proposed by \cite{zhang2019quaternion}. DualE \cite{cao2021dual} introduces translation operations and combines with quaternion rotations to encode additional relation patterns, including multiplicity. BiQUE \cite{guo2021bique} further includes hyperbolic rotations and scaling operations to better model hierarchical semantics. 3H-TH adds quaternion rotation to hyperbolic embedding model MuRP to further improve link prediction performance. 

Quite a few models also leverage scaling operation that is first demonstrated by PairRE \cite{chao-etal-2021-pairre} to have good performance in link prediction. Both LinearRE \cite{peng2020lineare} and TranSHER \cite{li2022transher} introduce translation vectors in the scaling-based scoring functions, but give different interpretations. LinearRE treats triple encoding as a linear regression problem, whereas TranSHER frames it as a translating distance model on a hyper-ellipsoid surface. Inspired by the idea of compounding affine operations for image manipulation, CompoundE \cite{ge2022compounde} further includes rotation operators to encode non-commutative relation compositions. Both ReflectE \cite{zhang2022knowledge} and HousE \cite{li2022house} encodes relations with householder reflections that have intrinsic symmetry property. ReflectE further explores the effect of combining translation vectors for also modeling antisymmetric relations in KG. On the other hand, HousE evaluates the effect of having a sequence of householder reflections on the link prediction performance. However, ATTH \cite{chami2020low} was in fact the first work that introduced reflection operations in KGE. Additional operators have also been applied to existing KGEs to encode temporal information. For instance, TeRo \cite{xu2020tero} applies 2D rotation to head and tail entity embedding in TransE to encode time-specific information in temporal KG quadruples. Similarly, RotateQVS \cite{chen2022rotateqvs} adds a time-specific translation vector to entity embedding in Rotate3D \cite{gao2020rotate3d}, which leverages quaternion rotation. CompoundE3D \cite{ge2023knowledge} gets inspired by the evolution from RotatE to Rotate3D and proposes a KGE that unifies all geometric operators including translation, scaling, rotation, reflection, and shear in 3D space. Apart from proposing a unified framework, CompoundE3D also suggests a beam search-based procedure for finding the optimal KGE design for each KG dataset.

\section{Unified Framework for Distance-based KGE with Affine Transformations: CompoundE and CompoundE3D}\label{sec:unified_framwork}
In this section, we will introduce the detailed formulation of CompoundE, followed by CompoundE3D. For both CompoundE and CompoundE3D, we have three forms of scoring functions, namely
\begin{itemize}
    \item CompoundE-Head
\begin{equation}
    f_r(h,t) = \|\mathbf{M_r\cdot h - t }\|,
\end{equation}
\item CompoundE-Tail
\begin{equation}
    f_r(h,t) = \|\mathbf{h - \hat{M}_r\cdot t }\|,
\end{equation}
\item CompoundE-Complete
\begin{equation}
    f_r(h,t) = \|\mathbf{M_r\cdot h - \hat{M}_r\cdot t }\|,
\end{equation}
\end{itemize}
where $\mathbf{M_r}$ and $\mathbf{\hat{M}_r}$ are geometric operators to be designed. We first discuss the CompoundE operators, which include translation, scaling, and 2D rotation operations.
\subsection{CompoundE Operators}
First, we think it is helpful to formally introduce some definitions in group theory.
\begin{defn}
The special orthogonal group is defined as
\end{defn}
\begin{equation}
\medmath{\mathbf{SO}(n) = \left \{ \mathbf{A} \bigg| \mathbf{A}\in \mathbf{GL}_n(\mathbb{R}),\mathbf{A}^{\top}\mathbf{A} = \mathbf{I}, \det(\mathbf{A})=1 \right \}}.
\end{equation}
\begin{defn}
The special Euclidean group is defined as
\end{defn}
\begin{equation}\label{eq:se}
\medmath{\mathbf{SE}(n) = \left \{ \mathbf{A} \bigg| \mathbf{A}=
    \begin{bmatrix}
      \mathbf{R} & \mathbf{v} \\
      \mathbf{0} & 1
    \end{bmatrix}, \mathbf{R}\in \mathbf{SO}(n), \mathbf{v}\in \mathbb{R}^n\right \}.}
\end{equation}
\begin{defn}
The affine group is defined as
\end{defn}
\begin{equation}\label{eq:aff}
    \medmath{\mathbf{Aff}(n) = \left \{ \mathbf{M} \bigg| \mathbf{M}=
    \begin{bmatrix}
      \mathbf{A} & \mathbf{v} \\
      \mathbf{0} & 1
    \end{bmatrix}, \mathbf{A}\in \mathbf{GL}_n(\mathbb{R}), \mathbf{v}\in 
    \mathbb{R}^n\right \}.}
\end{equation}
By comparing Eqs. (\ref{eq:se}) and (\ref{eq:aff}), we see that
$\mathbf{SE}(n)$ is a subset of $\mathbf{Aff}(n)$. 

Without loss of generality, consider $n=2$. If $\mathbf{M} \in \mathbf{Aff}(2)$, we have
\begin{equation}
    \mathbf{M} = 
  \begin{bmatrix}
    \mathbf{A} & \mathbf{v} \\
    \mathbf{0} & 1
  \end{bmatrix} , \mathbf{A}\in\mathbb{R}^{2\times2} , \mathbf{v}\in \mathbb{R}^{2}.
\end{equation}

The 2D translational matrix can be written as
\begin{equation}
    \mathbf{T} = \medmath{\begin{bmatrix}
    1 & 0 & v_x \\
    0 & 1 & v_y \\
    0 & 0 & 1
  \end{bmatrix}},
\end{equation}
while the 2D rotational matrix can be expressed as
\begin{equation}
    \mathbf{R} = \medmath{\begin{bmatrix}
    \cos(\theta) & -\sin(\theta) & 0 \\
    \sin(\theta) & \cos(\theta) & 0 \\
    0 & 0 & 1
  \end{bmatrix}.}
\end{equation}
It is easy to verify that they are both special Euclidean groups (i.e.
$\mathbf{T}\in \mathbf{SE}(2)$ and $\mathbf{R}\in \mathbf{SE}(2)$).  On the
other hand, the 2D scaling matrix is in form of
\begin{equation}
    \mathbf{S} = \medmath{\begin{bmatrix}
    s_x & 0 & 0 \\
    0 & s_y & 0 \\
    0 & 0 & 1
  \end{bmatrix}.}
\end{equation}
It is not a special Euclidean group but an affine group of $n=2$
(i.e., $\mathbf{S}\in\mathbf{Aff}(2)$).

Compounding translation and rotation operations, we can get a
transformation in the special Euclidean group,
\begin{equation}
\begin{aligned}
    \mathbf{T}\cdot \mathbf{R} &= \medmath{\begin{bmatrix}
    1 & 0 & v_x \\
    0 & 1 & v_y \\
    0 & 0 & 1
  \end{bmatrix}\begin{bmatrix}
    \cos(\theta) & -\sin(\theta) & 0 \\
    \sin(\theta) & \cos(\theta) & 0 \\
    0 & 0 & 1
  \end{bmatrix}} \\&= \medmath{\begin{bmatrix}
    \cos(\theta) & -\sin(\theta) & v_x \\
    \sin(\theta) & \cos(\theta) & v_y \\
    0 & 0 & 1
  \end{bmatrix}} \in \mathbf{SE}(2).
\end{aligned}
\end{equation}
Yet, if we add the scaling operation, the compound will belong to the
Affine group. One of such compound operator can be written as
\begin{equation}
\begin{aligned}
    \mathbf{T}\cdot \mathbf{R}\cdot \mathbf{S} &= \medmath{\begin{bmatrix}
    s_x\cos(\theta) & -s_y\sin(\theta) & v_x \\
    s_x\sin(\theta) & s_y\cos(\theta) & v_y \\
    0 & 0 & 1
  \end{bmatrix}} \in \mathbf{Aff}(2).
\end{aligned}
\end{equation}
When $s_x\neq0$ and $s_y\neq0$, the compound operator is invertible.
It can be written in form of
\begin{equation}
  \mathbf{M}^{-1} = 
  \begin{bmatrix}
    \mathbf{A}^{-1} & -\mathbf{A}^{-1}\mathbf{v} \\
    \mathbf{0} & 1
  \end{bmatrix}.
\end{equation}

In actual implementation, a high-dimensional relation operator can
be represented as a block diagonal matrix in the form of
\begin{equation}\label{eq:relation_operator_2d}
    \mathbf{M_r = \textbf{diag}(O_{r,1}, O_{r,2}, \dots, O_{r,n})},
\end{equation}
where $\mathbf{O_{r,i}}$ is the compound operator at the $i$-th stage. We can multiply $\mathbf{M_r\cdot v}$ in the following manner,

\begin{align}\label{large_matrix_2d}
    \setlength{\tabcolsep}{4pt}
    \centering
    \left[
    \begin{tabular}{cc|cc|cc|cc}
     $\mathbf{O}_{r,1}$ & {} & $0$ & {} & $\ldots$ & {} & $0$ & {} \\
     {} & {} & {} & {} & {} & {} & {} & {} \\
     \hline
     $0$ & {} & $\mathbf{O}_{r,2}$ & {} & $\ldots$ & {} & $0$ & {} \\
     {} & {} & {} & {} & {} & {} & {} & {} \\
     \hline
     $\vdots$ & {} & $\vdots$ & {} & $\ddots$ & {} & $\vdots$ & {} \\
     {} & {} & {} & {} & {} & {} & {} & {} \\
     \hline
     $0$ & {} & $0$ & {} & $\ldots$ & {} & $\mathbf{O}_{r,n}$ & {} \\
     {} & {} & {} & {} & {} & {} & {} & {} \\
    \end{tabular}
    \right]
    \left[
    \begin{tabular}{c}
         $x_1$ \\
         $y_1$ \\
         \hline
         $x_2$ \\
         $y_2$ \\
         \hline
         \\
         \vdots \\
         \hline
         $x_n$ \\
         $y_n$ 
    \end{tabular}
    \right]
\end{align}
where $\mathbf{v} = [x_1, y_1, x_2, y_2, \dots, x_n, y_n]^T$ are $2n$ dimensional entity vectors that are split into multiple 2d subspaces.

\subsection{CompoundE3D}


\begin{figure*}[t!]
    \centering
    \begin{subfigure}[t]{0.32\textwidth}
        \centering
        \includegraphics[height=1.6in]{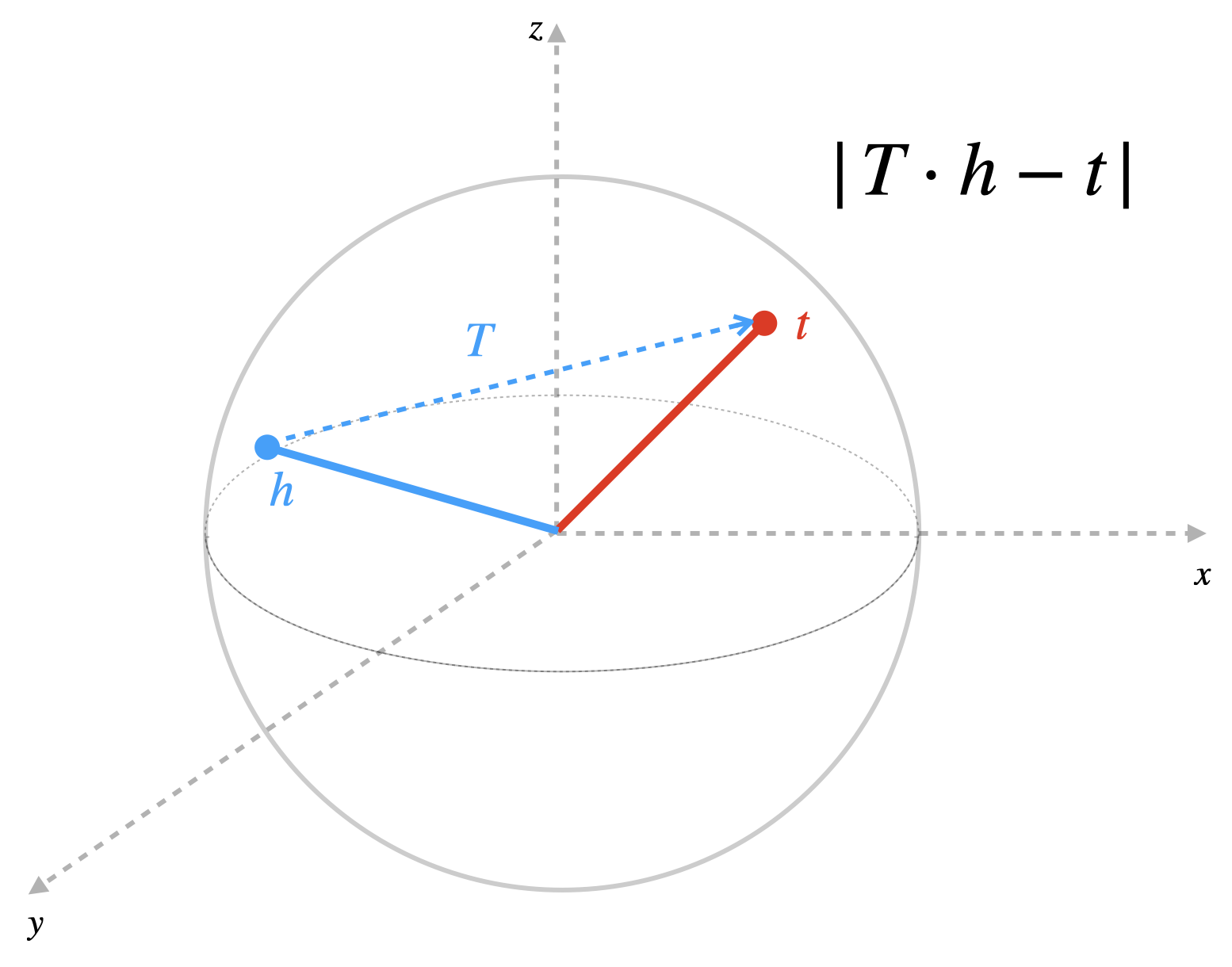}
        \caption{Translation}
        \label{fig:translate}
    \end{subfigure}%
    ~ 
    \begin{subfigure}[t]{0.32\textwidth}
        \centering
        \includegraphics[height=1.6in]{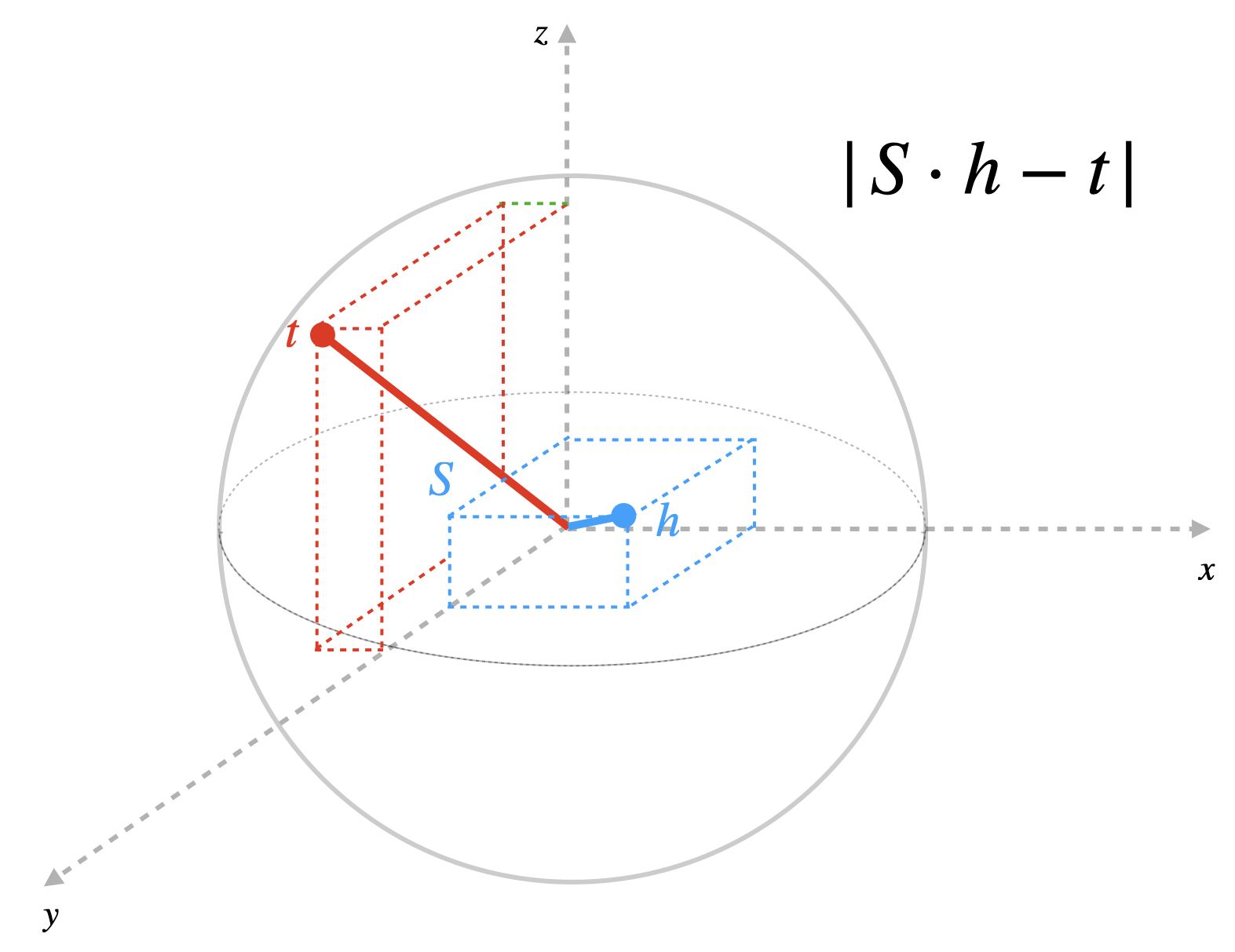}
        \caption{Scaling}
        \label{fig:scaling}
    \end{subfigure}
    ~
    \begin{subfigure}[t]{0.32\textwidth}
    \centering
    \includegraphics[height=1.6in]{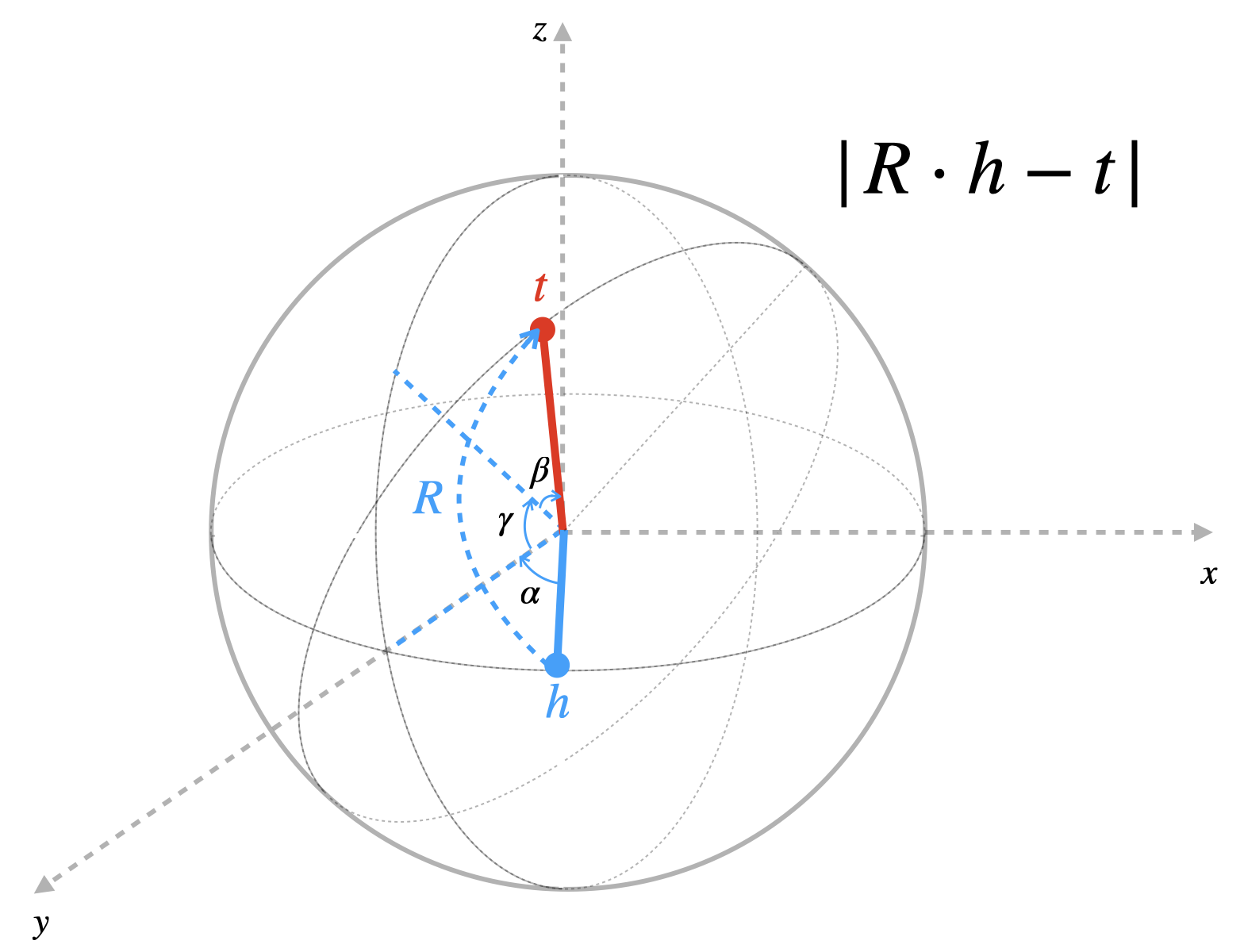}
    \caption{Rotation}
    \label{fig:rotation}
    \end{subfigure}
    
    \begin{subfigure}[t]{0.32\textwidth}
        \centering
        \includegraphics[height=1.6in]{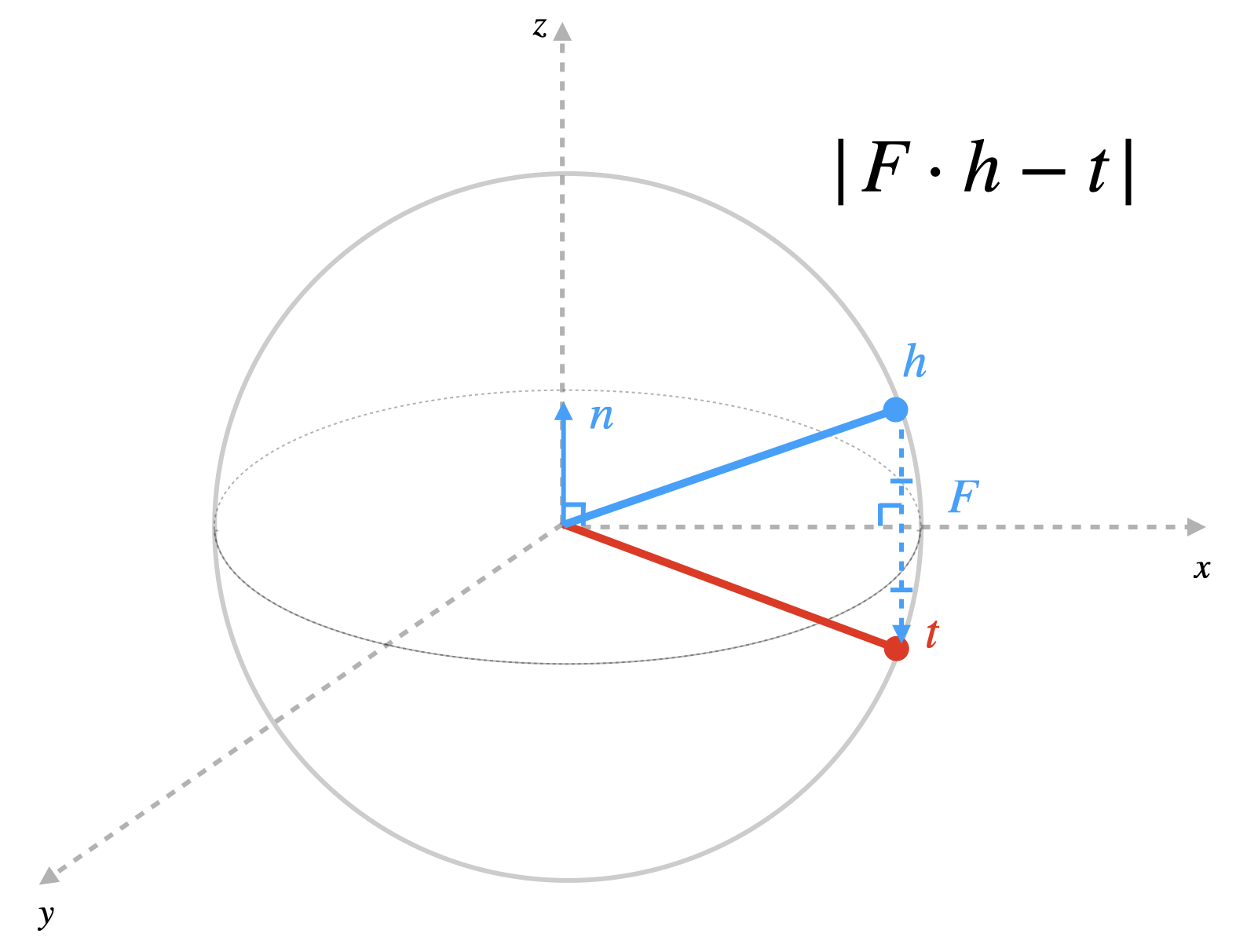}
        \caption{Reflection}
        \label{fig:reflection}
    \end{subfigure}%
    ~ 
    \begin{subfigure}[t]{0.32\textwidth}
        \centering
        \includegraphics[height=1.6in]{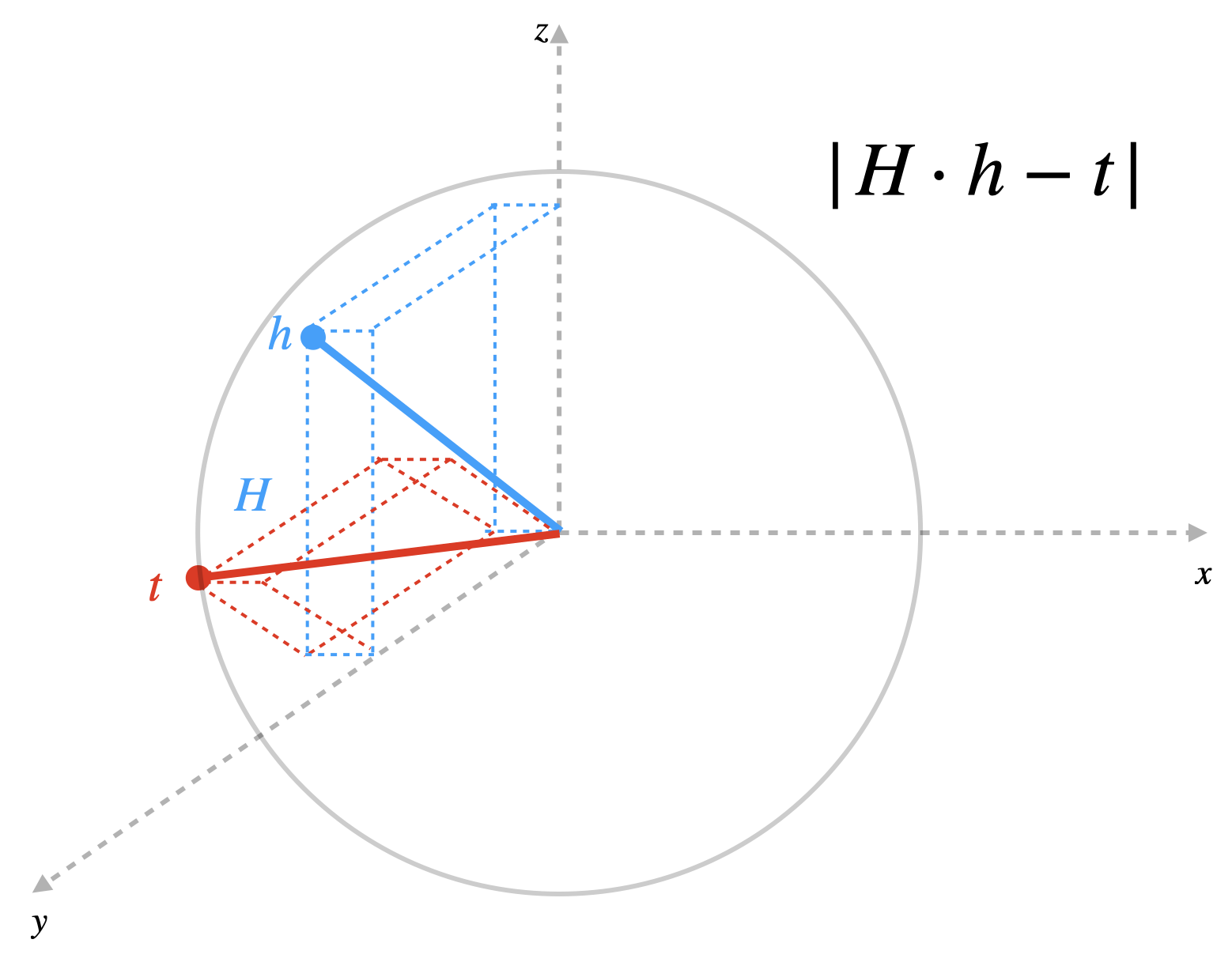}
        \caption{Shear}
        \label{fig:shear}
    \end{subfigure}
    ~
    \begin{subfigure}[t]{0.32\textwidth}
    \centering
    \includegraphics[height=1.6in]{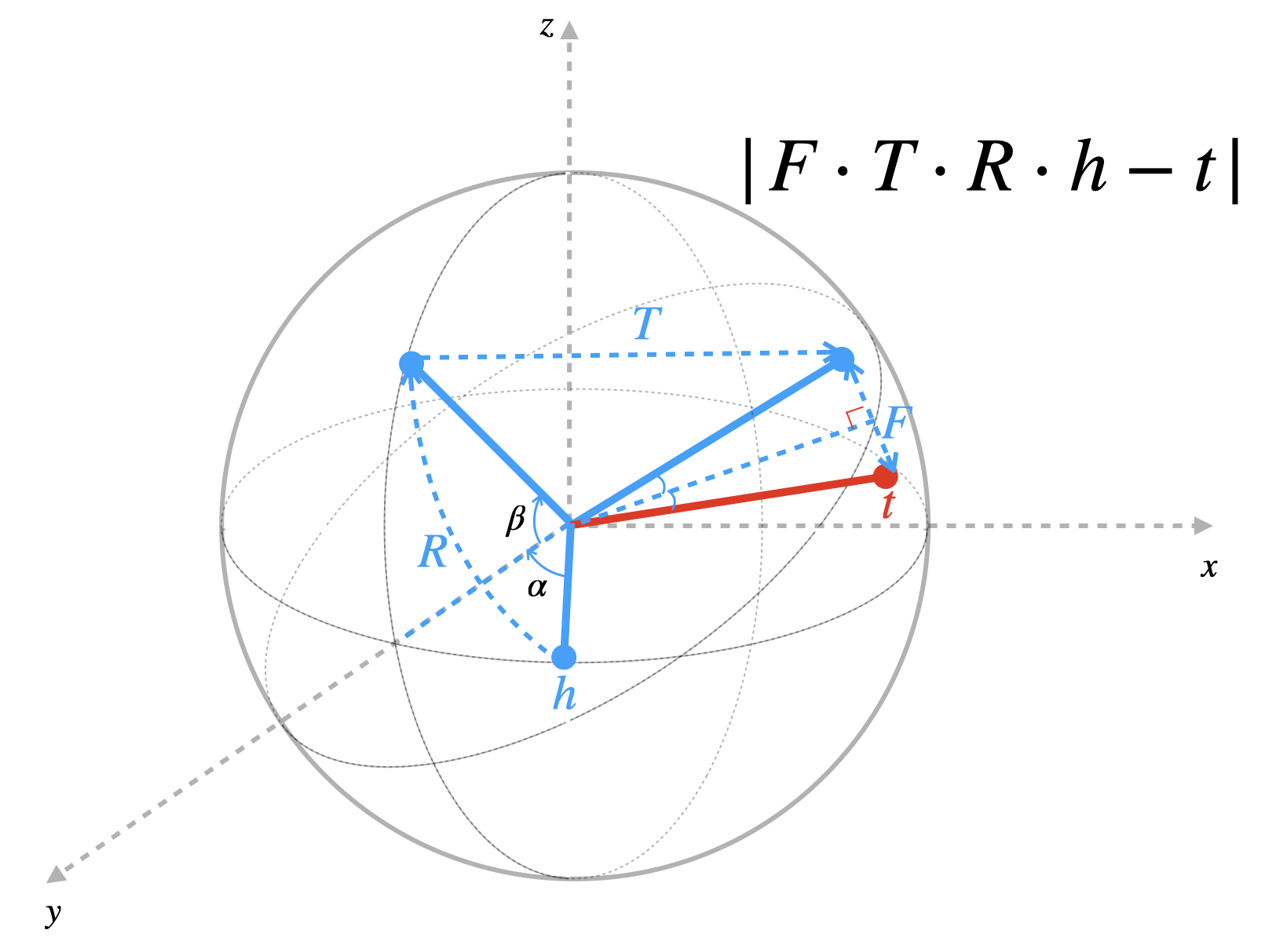}
    \caption{Compound}
    \label{fig:compound}
    \end{subfigure}
\caption{Composing different geometric operations in the 3D subspace.}\label{fig:diff_operation}
\end{figure*}

\subsubsection{Translation} Component $\mathbf{T}\in\mathbf{SE}(3)$,
illustrated by Fig. \ref{fig:translate}, is defined as
\begin{equation}
    \mathbf{T} = \begin{bmatrix}
    1 & 0 & 0 & v_x \\
    0 & 1 & 0 & v_y \\
    0 & 0 & 1 & v_z \\
    0 & 0 & 0 & 1
  \end{bmatrix},
\end{equation}

\subsubsection{Scaling} Component $\mathbf{S}\in\mathbf{Aff}(3)$,
illustrated by Fig. \ref{fig:scaling}, is defined as
\begin{equation}
    \mathbf{S} = \begin{bmatrix}
    s_x & 0 & 0 & 0 \\
    0 & s_y & 0 & 0 \\
    0 & 0 & s_z & 0 \\
    0 & 0 & 0 & 1
  \end{bmatrix},
\end{equation}

\subsubsection{Rotation} Component $\mathbf{R}\in\mathbf{SO}(3)$,
illustrated by Fig. \ref{fig:rotation}, is defined as
\begin{equation}
    \mathbf{R} = \mathbf{R}_z(\alpha)\mathbf{R}_y(\beta)\mathbf{R}_x(\gamma) = 
    \begin{bmatrix}
    a & b & c & 0 \\
    d & e & f & 0 \\
    g & h & i & 0 \\
    0 & 0 & 0 & 1
  \end{bmatrix},
\end{equation}
where
\begin{equation}
    \begin{aligned}
        a &= \cos(\alpha)\cos(\beta), \\
        b &= \cos(\alpha)\sin(\beta)\sin(\gamma)-\sin(\alpha)\cos(\gamma), \\
        c &= \cos(\alpha)\sin(\beta)\cos(\gamma)+\sin(\alpha)\sin(\gamma), \\
        d &= \sin(\alpha)\cos(\beta), \\
        e &= \sin(\alpha)\sin(\beta)\sin(\gamma)+\cos(\alpha)\cos(\gamma), \\
        f &= \sin(\alpha)\sin(\beta)\cos(\gamma)-\cos(\alpha)\sin(\gamma), \\
        g &= -\sin(\beta), \\
        h &= \cos(\beta)\sin(\gamma), \\
        i &= \cos(\beta)\cos(\gamma). \\
    \end{aligned}
\end{equation}
This general 3D rotation operator is the result of compounding yaw,
pitch, and roll rotations. They are, respectively, defined as
\begin{itemize}
\item Yaw rotation component:
\begin{equation}
    \mathbf{R}_z(\alpha) = \begin{bmatrix}
    \cos(\alpha) & -\sin(\alpha) & 0 & 0 \\
    \sin(\alpha) & \cos(\alpha) & 0 & 0 \\
    0 & 0 & 1 & 0 \\
    0 & 0 & 0 & 1
  \end{bmatrix},
\end{equation}
\item Pitch rotation component:
\begin{equation}
    \mathbf{R}_y(\beta) = \begin{bmatrix}
    \cos(\beta) & 0 & -\sin(\beta) & 0 \\
    0 & 1 & 0 & 0 \\
    \sin(\beta) & 0 & \cos(\beta) & 0 \\
    0 & 0 & 0 & 1
  \end{bmatrix},
\end{equation}
\item Roll rotation component:
\begin{equation}
    \mathbf{R}_x(\gamma) = \begin{bmatrix}
    1 & 0 & 0 & 0 \\
    0 & \cos(\gamma) & -\sin(\gamma) & 0 \\
    0 & \sin(\gamma) & \cos(\gamma) & 0 \\
    0 & 0 & 0 & 1
  \end{bmatrix}.
\end{equation}
\end{itemize}

\subsubsection{Reflection} Component $\mathbf{F}\in\mathbf{SO}(3)$,
illustrated by Fig. \ref{fig:reflection}, is defined as
\begin{equation}
    \mathbf{F} = \begin{bmatrix} 
    1 - 2n_x^2 & -2n_x n_y & -2n_x n_z & 0 \\
    -2n_x n_y & 1 - 2 n_y^2 & -2n_y n_z & 0 \\
    -2n_x n_z & -2n_y n_z & 1 - 2n_z^2 & 0 \\
    0 & 0 & 0 & 1 
    \end{bmatrix}.
\end{equation}
The above expression is derived from the Householder reflection, $\mathbf{F =
I - 2nn^T}$. In the 3D space, $\mathbf{n}$ is a 3-D unit vector that is
perpendicular to the reflecting hyper-plane, $\mathbf{n} = [n_x, n_y,
n_z]$. 

\subsubsection{Shear} Component $\mathbf{H}\in\mathbf{Aff}(3)$,
illustrated by Fig. \ref{fig:shear}, is defined as
\begin{equation}
    \mathbf{H} = \mathbf{H}_{yz} \mathbf{H}_{xz} \mathbf{H}_{xy} = \begin{bmatrix}
    1 & \text{Sh}^y_x & \text{Sh}^z_x & 0 \\
    \text{Sh}^x_y & 1 & \text{Sh}^z_y & 0 \\
    \text{Sh}^x_z & \text{Sh}^y_z & 1 & 0 \\
    0 & 0 & 0 & 1
  \end{bmatrix}.
\end{equation}
The shear operator is the result of compounding 3 operators:
$\mathbf{H}_{yz}$, $\mathbf{H}_{xz}$, and $\mathbf{H}_{xy}$ 
They are mathematically defined as
\begin{eqnarray}
\mathbf{H}_{yz} & = & \begin{bmatrix}
    1 & 0 & 0 & 0 \\
    \text{Sh}^x_y & 1 & 0 & 0 \\
    \text{Sh}^x_z & 0 & 1 & 0 \\
    0 & 0 & 0 & 1 \\
\end{bmatrix}, \\
\mathbf{H}_{xz} & = & \begin{bmatrix}
    1 & \text{Sh}^y_x & 0 & 0 \\
    0 & 1 & 0 & 0 \\
    0 & \text{Sh}^y_z & 1 & 0 \\
    0 & 0 & 0 & 1 \\
\end{bmatrix}, \\
\mathbf{H}_{xy} & = & \begin{bmatrix}
    1 & 0 & \text{Sh}^z_x & 0 \\
    0 & 1 & \text{Sh}^z_y & 0 \\
    0 & 0 & 1 & 0 \\
    0 & 0 & 0 & 1 \\
\end{bmatrix}.
\end{eqnarray}
Matrix $\mathbf{H}_{yz}$ has a physical meaning - the shear
transformation that shifts the $y$- and $z$- components by a factor of
the $x$ component.  Similar physical interpretations are applied to
$\mathbf{H}_{xz}$ and $\mathbf{H}_{xy}$. 

The above transformations can be cascaded to yield a compound operator; e.g.,
\begin{equation}
\mathbf{O = T\cdot S\cdot R\cdot F\cdot H},
\end{equation}

In the actual implementation, we use the operator's representation in regular Cartesian coordinates instead of the homogeneous coordinate. Furthermore, a high-dimensional relation operator can
be represented as a block diagonal matrix in the form of
\begin{equation}\label{eq:relation_operator}
    \mathbf{M_r = \textbf{diag}(O_{r,1}, O_{r,2}, \dots, O_{r,n})},
\end{equation}
where $\mathbf{O_{r,i}}$ is the compound operator at the $i$-th stage. We can multiply $\mathbf{M_r\cdot v}$ in the following manner,

\begin{align}\label{large_matrix}
    \setlength{\tabcolsep}{4pt}
    \centering
    \left[
    \begin{tabular}{ccc|ccc|ccc|ccc}
     {} & {} & {} & {} & {} & {} & {} & {} & {} & {} & {} & {} \\
     {} & $\mathbf{O}_{r,1}$ & {} & {} & $0$ & {} & {} & $\ldots$ & {} & {} & $0$ & {} \\
     {} & {} & {} & {} & {} & {} & {} & {} & {} & {} & {} & {} \\
     \hline
     {} & {} & {} & {} & {} & {} & {} & {} & {} & {} & {} & {} \\
     {} & $0$ & {} & {} & $\mathbf{O}_{r,2}$ & {} & {} & $\ldots$ & {} & {} & $0$ & {} \\
     {} & {} & {} & {} & {} & {} & {} & {} & {} & {} & {} & {} \\
     \hline
     {} & {} & {} & {} & {} & {} & {} & {} & {} & {} & {} & {} \\
     {} & $\vdots$ & {} & {} & $\vdots$ & {} & {} & $\ddots$ & {} & {} & $\vdots$ & {} \\
     {} & {} & {} & {} & {} & {} & {} & {} & {} & {} & {} & {} \\
     \hline
     {} & {} & {} & {} & {} & {} & {} & {} & {} & {} & {} & {} \\
     {} & $0$ & {} & {} & $0$ & {} & {} & $\ldots$ & {} & {} & $\mathbf{O}_{r,n}$ & {} \\
     {} & {} & {} & {} & {} & {} & {} & {} & {} & {} & {} & {} 
    \end{tabular}
    \right]
    \left[
    \begin{tabular}{c}
         $x_1$ \\
         $y_1$ \\
         $z_1$ \\
         \hline
         $x_2$ \\
         $y_2$ \\
         $z_2$ \\
         \hline
         \\
         \vdots \\
         \\
         \hline
         $x_n$ \\
         $y_n$ \\
         $z_n$
    \end{tabular}
    \right]
\end{align}
where $\mathbf{v} = [x_1, y_1, z_1, x_2, y_2, z_2, \dots, x_n, y_n, z_n]^T$ are $3n$ dimensional entity vectors that are split into multiple 3d subspaces.

\section{Dataset and Evaluation}\label{sec:datasets_and_evaluation}

\subsection{Open Knowledge Graphs and Benchmarking Datasets}

Table \ref{tab:open_knowledge_graph} collects a set of KG that are available for public access. We list these KGs according to the time of their first release. We provide information for the number of entities, the number of facts to indicate the scale and size of each KG. Among these public KGs, WordNet \cite{miller1995wordnet} is the oldest and it is first created as a lexical database of English words. Words are connected with semantic relations including Synonymy, Antonymy, Hyponymy, Meronymy, Troponomy, and Entailment. There are similarities and differences between the KGs. For example, both DBpedia \cite{auer2007dbpedia} and YAGO \cite{suchanek2007yago} are derived from information in Wikipedia infoboxes. However, YAGO also includes information from GeoNames that contains spatial information. In addition, DBpedia, YAGO, and Wikidata \cite{vrandevcic2014wikidata} contain multilingual information. Both ConceptNet \cite{speer2017conceptnet} and OpenCyc contain a wide range of commonsense concepts and relations.

Table \ref{tab:kg_dataset_lp} contains commonly used benchmarking datasets. These datasets have different sizes. Each of them covers a different domain. For example, UMLS contains biomedical concepts from the Unified Medical Language System. Similarly, OGB-biokg is also a biomedical KG but with a larger size. Kinship contains relationship between members of a tribe and Countries contains relationships between countries and regions. As indicated by the dataset names, many of them are subsets of the public KGs that are described above. CoDEx is extracted from Wikidata, but presents 3 different versions and each has different degrees and densities. Among these datasets, FB15K-237 and WN18RR are the most adopted datasets for performance benchmarking since inverse relations are removed to avoid the test leakage problem.


\begin{longtable}{c|c|c|c|p{2.5cm}|p{2.5cm}|p{4cm}}
\caption{Major Open Knowledge Graphs.} \label{tab:open_knowledge_graph}\\
      \hline
      \textbf{Name} & \textbf{Year} & \textbf{\# Entities} & \textbf{\# Facts} & \textbf{Source} & \textbf{Highlights} & \textbf{Access Link}\\
      \hline     
      WordNet & 1980s & 155K & 207K & Curated by experts & Lexical database of semantic relations between words & \url{https://wordnet.princeton.edu} \\
      \hline
      ConceptNet & 1999 & 44M & 2.4B & Crowdsourced human data and structured feeds & Commonsense concepts and relations & \url{https://conceptnet.io} \\
      \hline
      Freebase & 2007 & 44M & 2.4B & Crowdsourced human data and structured feeds & One of the first public KBs/KGs & \url{https://developers.google.com/freebase/} \\
      \hline 
      DBpedia & 2007 & 4.3M & 70M & Automatically extracted structured data from Wikipedia & Multilingual, Cross-Domain & \url{https://dbpedia.org/sparql} \\
      \hline  
      YAGO & 2008 & 10M & 120M & Derived from Wikipedia, WordNet, and GeoNames & Multilingual, Fine grained entity types & \url{https://yago-knowledge.org/sparql} \\
      \hline  
      Wikidata & 2012 & 50M & 500M & Crowdsourced human curation & Collaborative, Multilingual, Structured & \url{https://query.wikidata.org} \\
      \hline 
      OpenCyc & 2001 & 2.4M & 240k & Created by domain experts & Commonsense concepts and relations & \url{http://www.cyc.com} \\
      \hline 
      NELL & 2010 & - & 50M & Extracted from Clueweb09 corpus (1B web pages) &  Each fact is given a confidence score. 2.8M beliefs has high confidence score. & \url{http://rtw.ml.cmu.edu/rtw/} \\
      \hline 
\end{longtable}


\begin{longtable}{c|cc|ccc|p{1cm}|p{4cm}}
\caption{Knowledge Graph Benchmarking Datasets for Link Prediction.} \label{tab:kg_dataset_lp}\\
      \hline
      \multirow{2}{*}{\textbf{Dataset}} & \multicolumn{6}{c|}{\textbf{Statistics}}  & \multirow{2}{*}{\textbf{Remarks}} \\\cline{2-7} 
      & \textbf{\#Ent} & \textbf{\#Rel} & \textbf{\#Train} & \textbf{\#Valid} & \textbf{\#Test} & \textbf{Avg. Deg.}  & \\
      \hline
      Kinship & 104 & 26 & 8,544 & 1,068 & 1,074 & 82.15 & Information about complex relational structure among members of a tribe.\\
      \hline
      UMLS & 135 & 49 & 5,216 & 652 & 661 & 38.63 & Biomedical relationships between categorized concepts of the Unified Medical Language System.\\
      \hline
      Countries & 272 & 2 & 1,111 & 24 & 24 & 4.35 & Relationships between countries, regions, and subregions.\\
      \hline
      FB15K & 14,951 & 1,345 & 483,142 & 50,000 & 59,071 & 13.2 & A subset of the Freebase. Textual descriptions of entities are available.\\
      \hline
      FB15K-237 & 14,951 & 237 & 272,115 & 17,535 & 20,466 & 19.74 & Derived from FB-15K by removing inverse relations to avoid test leakage problem.\\
      \hline
      WN18 & 40,943 & 18 & 141,442 & 5,000 & 5,000 & 1.2 & A subset of the WordNet. \\
      \hline
      WN18RR & 40,943 & 11 & 86,835 & 3,034 & 3,134 & 2.19 & Derived from WN18 by removing inverse relations to avoid test leakage problem.\\
      \hline
      YAGO3-10 & 123,182 & 37 & 1,079,040 & 5,000 & 5,000 & 9.6 & Subset of YAGO3 (extension of YAGO) that contains entities associated with at least 10 different relations, describing citizenship, gender, and profession of people.\\
      \hline
      DB100K & 99,604 & 470 & 597,572 & 50,000 & 50,000 & 12 & A subset of the DBpedia. Each entity appears in at least 20 different relations \\
      \hline
      CoDEx-S & 2,034 & 42 & 32,888 & 1,827 & 1,828 & 21.47 & Extracted from Wikidata. Each entity has degree at least 15. Hard negative samples are provided. \\
      \hline
      CoDEx-M & 17,050 & 51 & 185,584 & 10,310 & 10,311 & 13.45 & Extracted from Wikidata. Each entity has degree at least 10. Hard negative samples are provided. \\
      \hline
      CoDEx-L & 77,951 & 69 & 551,193 & 30,622 & 30,622 & 25.62 & Extracted from Wikidata. Each entity has degree at least 5. \\
      \hline
      OGB-Wikikg2 & 2,500,604 & 535 & 16,109k & 429k & 598k & 8.79 & Extracted from Wikidata. Triple split based on timestamp rather than random split. \\
      \hline
      OGB-Biokg & 93,773 & 51 & 4,763k & 163k & 163k & 47.5 & Created from a large number of biomedical data repositories. Contains information about diseases, proteins, drugs, side effects, and protein functions. \\
      \hline
\end{longtable}


\subsection{Evaluation Metrics and Leaderboard}

The link prediction performance of KGE models is typically evaluated using the Mean Reciprocal Rank (MRR) and Hits@$k$ metrics. The MRR is the average of the reciprocal ranks of the ground truth entities. The Hits@$k$ is the fraction of test triples for which the ground truth entity is ranked among the top $k$ candidates. The MRR and Hits@$k$ metrics are defined as follows:
\begin{itemize}
    \item The MRR is calculated as:
    \begin{equation}
        \text{MRR} = \frac{1}{|D|} \sum_{i \in D} \frac{1}{\text{Rank}_i},
    \end{equation}
    where $|D|$ is the number of test triples, and $\text{Rank}_i$ is the rank of the ground truth entity in the list of top candidates for the $i$th test triple.
    \item The Hits@k is calculated as:
    \begin{equation}
        \text{Hits}@k = \frac{1}{|D|} \sum_{i \in D} \mathds{1}\{\text{Rank}_i \leq k\},
    \end{equation}
    where $\mathds{1}\{\cdot\}$ is the indicator function.
\end{itemize}

Higher MRR and Hits@$k$ values indicate better model performance. This is because they mean that the model is more likely to rank the ground truth entity higher in the list of top candidates, and to rank it among the top $k$ candidates, respectively. In order to prevent the model from simply memorizing the triples in the KG and ranking them higher, the filtered rank is typically used to evaluate the link prediction performance of KGE models. The filtered rank is the rank of the ground truth entity in the list of top candidates, but only considering candidates that would result in unseen triples. In addition to the MRR and Hits@$k$ metrics, other evaluation metrics for KG completion include Mean Average Precision (MAP), Precision@$k$, etc. The choice of evaluation metric depends on the specific application. For example, if the goal is to rank the top entities for a given triple, then the MRR or Hits@$k$ metrics may be more appropriate. If the goal is to find all entities that are related to a given entity, then the MAP or Precision@$k$ metrics may be more appropriate.

We also show a comparison of the Hits@10 performance of recently published works for link prediction in Fig. \ref{fig:hit10_lb}. The figure includes the results on the FB15k, FB15k-237, WN18, WN18RR, and YAGO3-10 datasets. While KGE models such as MEIM \cite{tran2022meim} remain competitive, many of the pretrained language model (PLM) approaches such as SimKGC \cite{wang2022simkgc} and LMKE \cite{wang2022language} top the leaderboards. In the next section, we will discuss this emerging trend of using PLM for KG completion.

\begin{figure}[ht!]
    \centering
    \subfloat[FB15K]{\includegraphics[width=\textwidth]{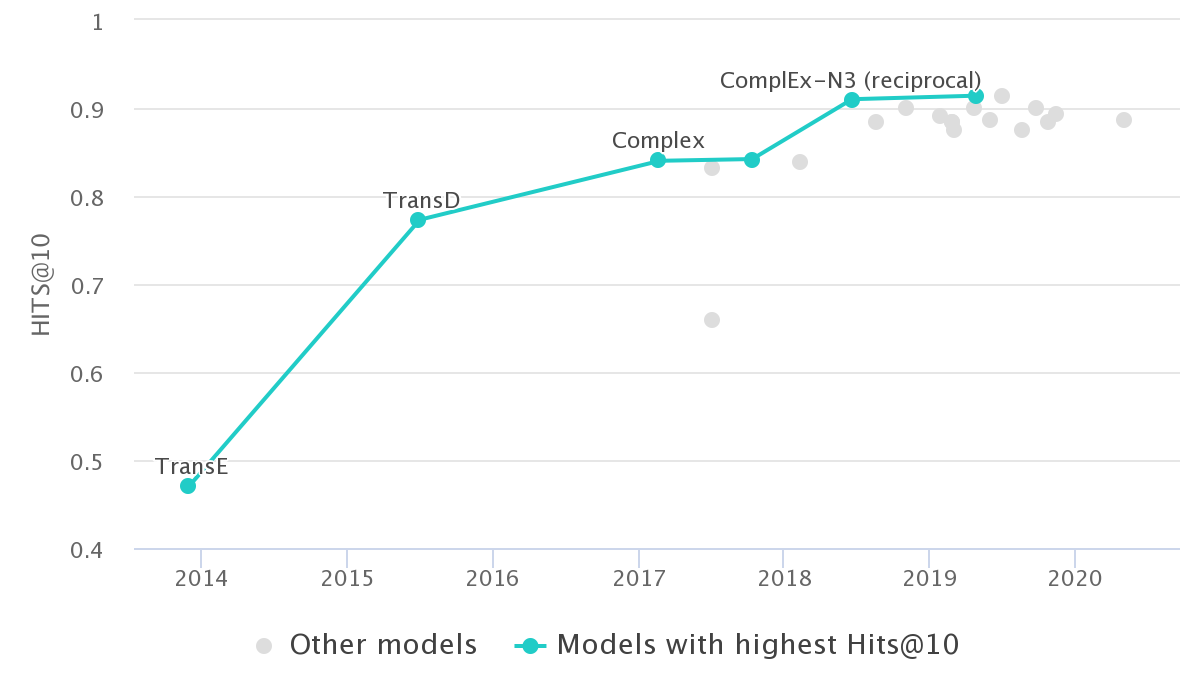}} \\
    \subfloat[FB15K-237]{\includegraphics[width=\textwidth]{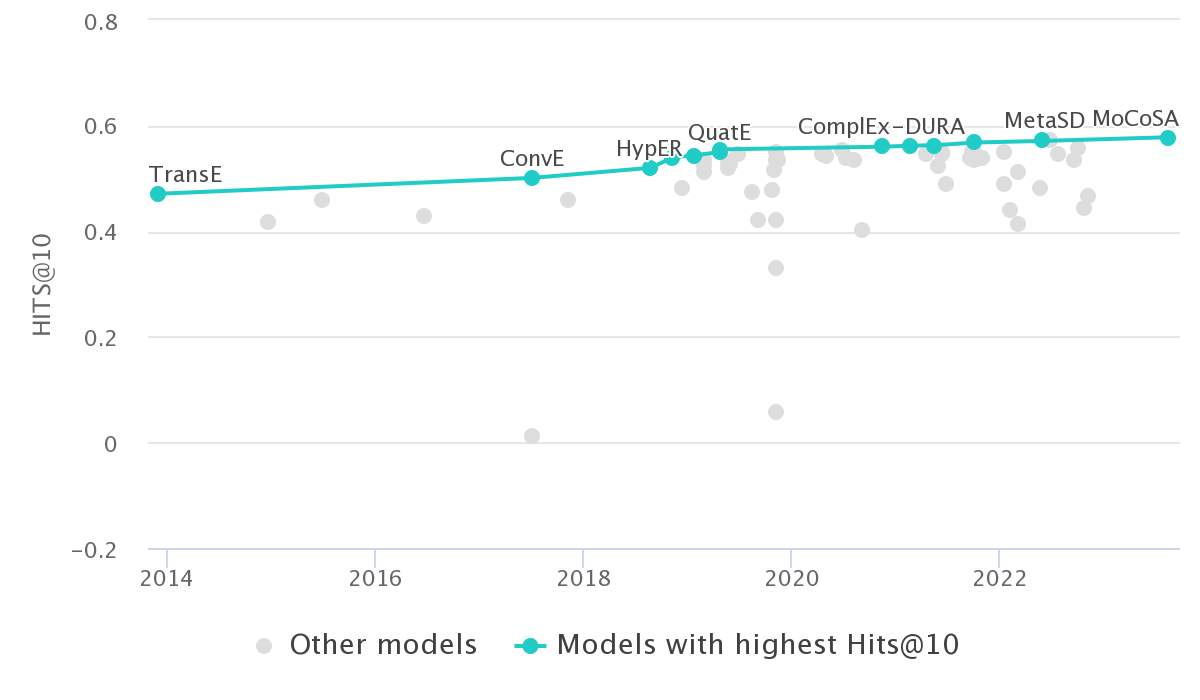}} \\
    \label{fig:wikikg_hyper_search}
    \caption{HITS@10 score of previous KGE models for datasets.\\\hspace{\textwidth} Source: \protect\url{https://paperswithcode.com/sota}}
\end{figure}

\begin{figure}[ht!]\ContinuedFloat 
    \subfloat[WN18]{\includegraphics[width=\textwidth]{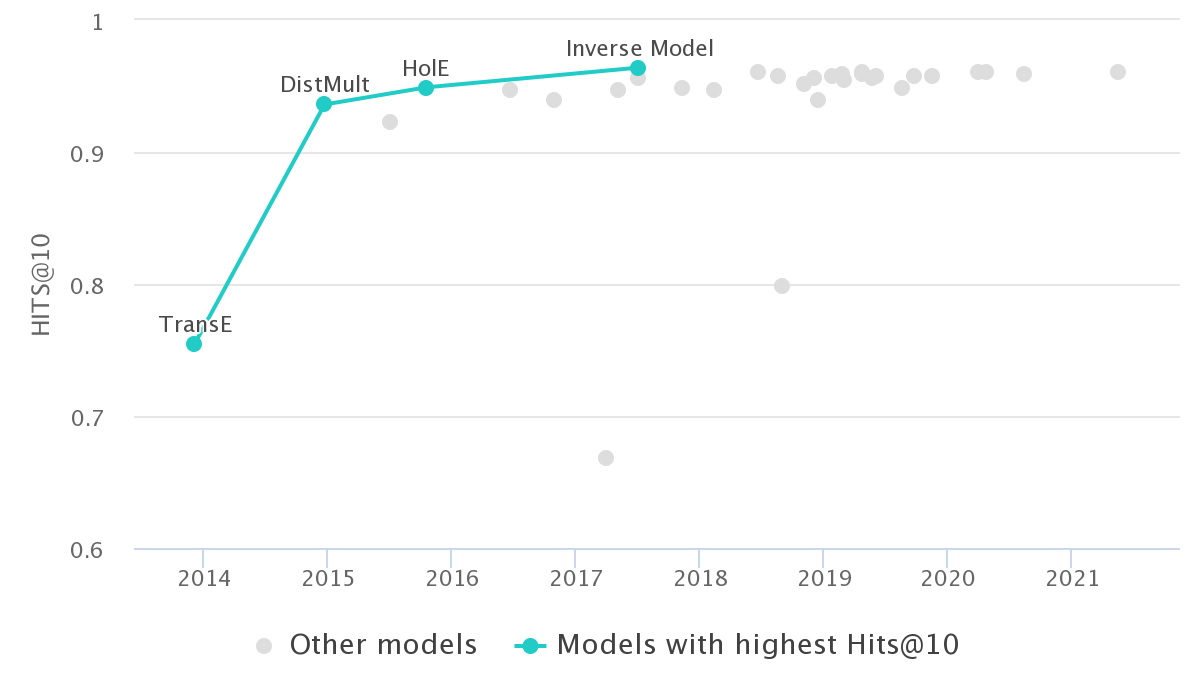}} \\
    \subfloat[WN18RR]{\includegraphics[width=\textwidth]{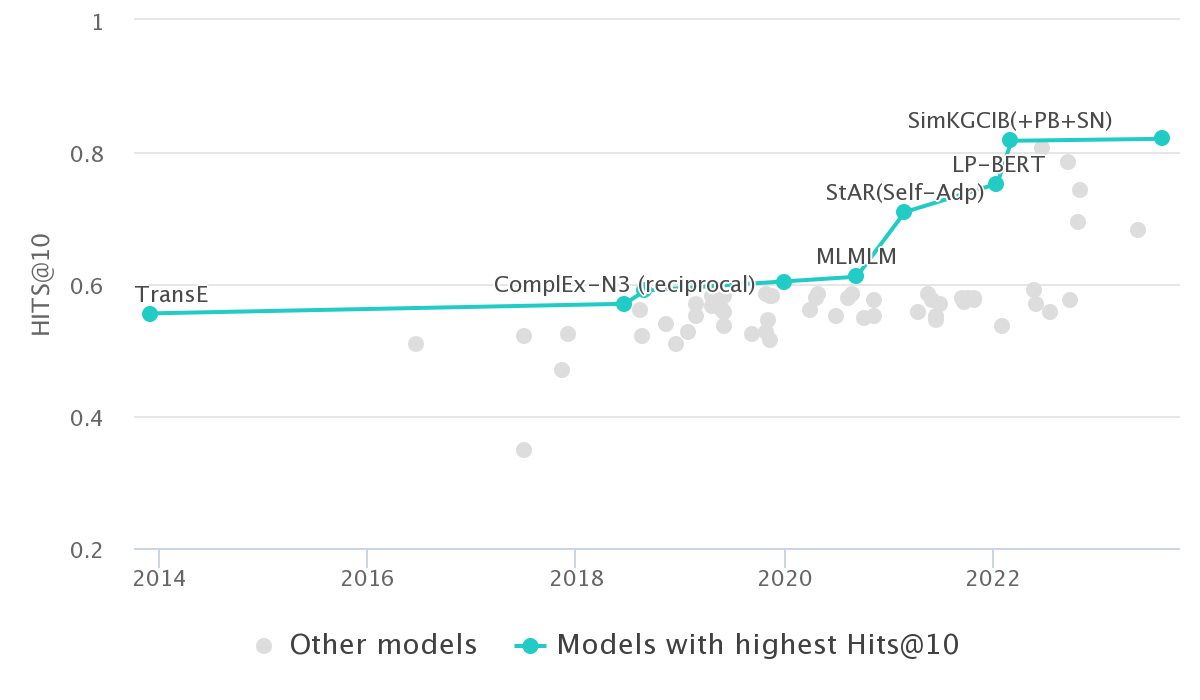}} \\
    \caption{HITS@10 score of previous KGE models for datasets (cont.). \\\hspace{\textwidth} Source: \protect\url{https://paperswithcode.com/sota}}
\end{figure}

\begin{figure}[ht!]\ContinuedFloat 
    \subfloat[YAGO3-10]{\includegraphics[width=\textwidth]{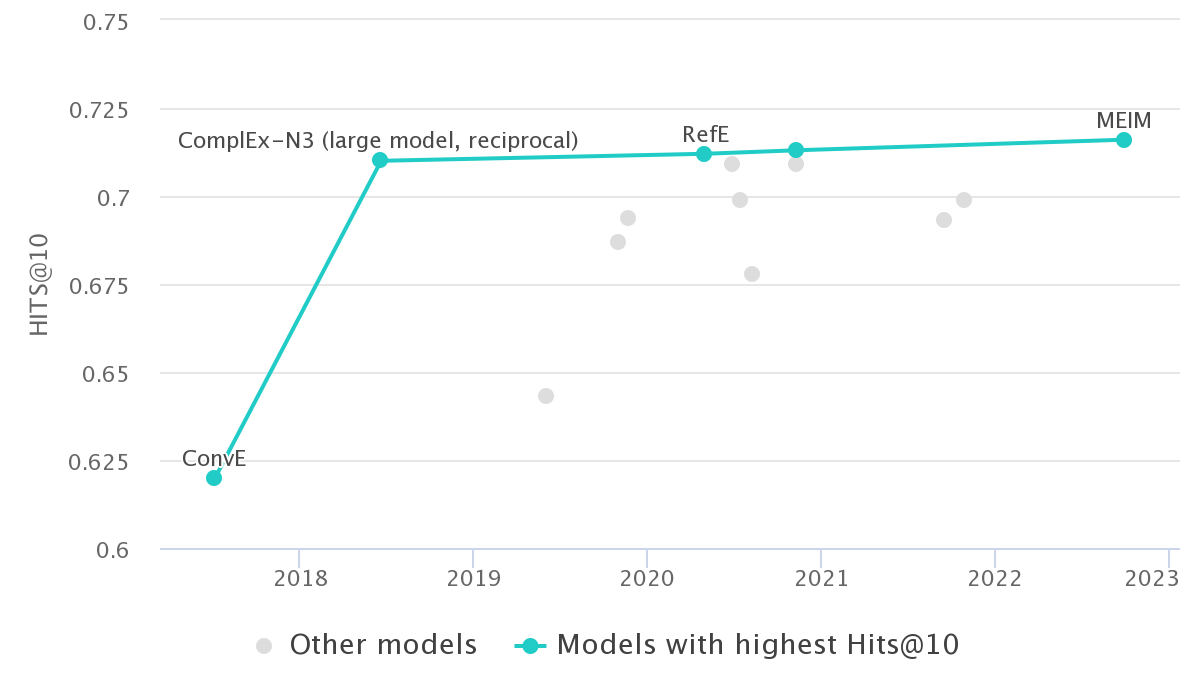}} \\
    \caption{HITS@10 score of previous KGE models for datasets (cont.) \\\hspace{\textwidth} Source: \protect\url{https://paperswithcode.com/sota}}\label{fig:hit10_lb}
\end{figure}

\section{Emerging Direction}\label{sec:emerging}

\subsection{Neural Network Models for Knowledge Graph Completion}
Before discussing the PLM approach, it is worthwhile to introduce neural network models for KG completion since PLMs also belong to this line of approach and the logic for training and inference between these models are similar. a Multilayer Perceptron (MLP) \cite{dong2014knowledge} is used to measure the likelihood of unseen triples for link prediction. NTN \cite{socher2013reasoning} adopts a bilinear tensor neural layer to model interactions between entities and relations of triples. ConvE \cite{dettmers2018convolutional} reshapes and stacks the head entity and the relation vector to form a 2D shape data, applies Convolutional Neural Networks (CNNs) to extract features, and uses extracted features to interact with tail embedding. R-GCN \cite{schlichtkrull2018modeling} applies a Graph Convolutional Network (GCN) and considers the neighborhood of each entity equally. CompGCN \cite{vashishth2020compositionbased} performs a composition operation over each edge in the neighborhood of a central node. The composed embeddings are then convolved with specific filters representing the original and the inverse relations, respectively. KBGAN \cite{cai2018kbgan} optimizes a generative adversarial network to generate the negative samples. KBGAT \cite{nathani2019learning} applies graph attention networks to capture both entity and relation features in any given entity’s neighborhood. ConvKB \cite{dai2018novel} applies 1D convolution on stacked entity and relation embeddings of a triple to extract feature maps and applies a nonlinear classifier to predict the likelihood of the triple. Structure-Aware Convolutional Network (SACN) \cite{shang2019end} uses a weighted-GCN encoder and a Conv-TransE decoder to extract the embedding. This synergy successfully leverages graph connectivity structure information. InteractE \cite{vashishth2020interacte} uses network design ideas including feature permutation, a novel feature reshaping, and circular convolution compared to ConvE and outperforms baseline models significantly. ParamE \cite{che2020parame} uses neural networks instead of relation embedding to model the relational interaction between head and tail entities. MLP, CNN, and gated structure layers are experimented and the gated layer turns out to be far more effective than embedding approaches. ReInceptionE \cite{xie2020reinceptione} applies Inception network to increase the interactions between head and relation embeddings. Relation-aware attention mechanism in the model aggregates the local neighborhood features and the global entity information. M-DCN \cite{zhang2020multi} adopts a multi-scale dynamic convolutional network to model complex relations such as 1-N, N-1, and N-N relations. A related model called KGBoost is a tree classifier-based method that proposes a novel negative sampling method and uses the XGBoost classifier for link prediction. GreenKGC \cite{wang2022greenkgc} is a modularized KGC method inspired by Discriminant Feature Learning (DFT) \cite{kuo2022green, yang2022supervised}, which extracts the most discriminative feature from trained embeddings for binary classifier learning.

\subsection{Pretrained Language Models for Knowledge Graph Completion}
\begin{figure}[ht!]
\centering
\includegraphics[width=1\columnwidth]{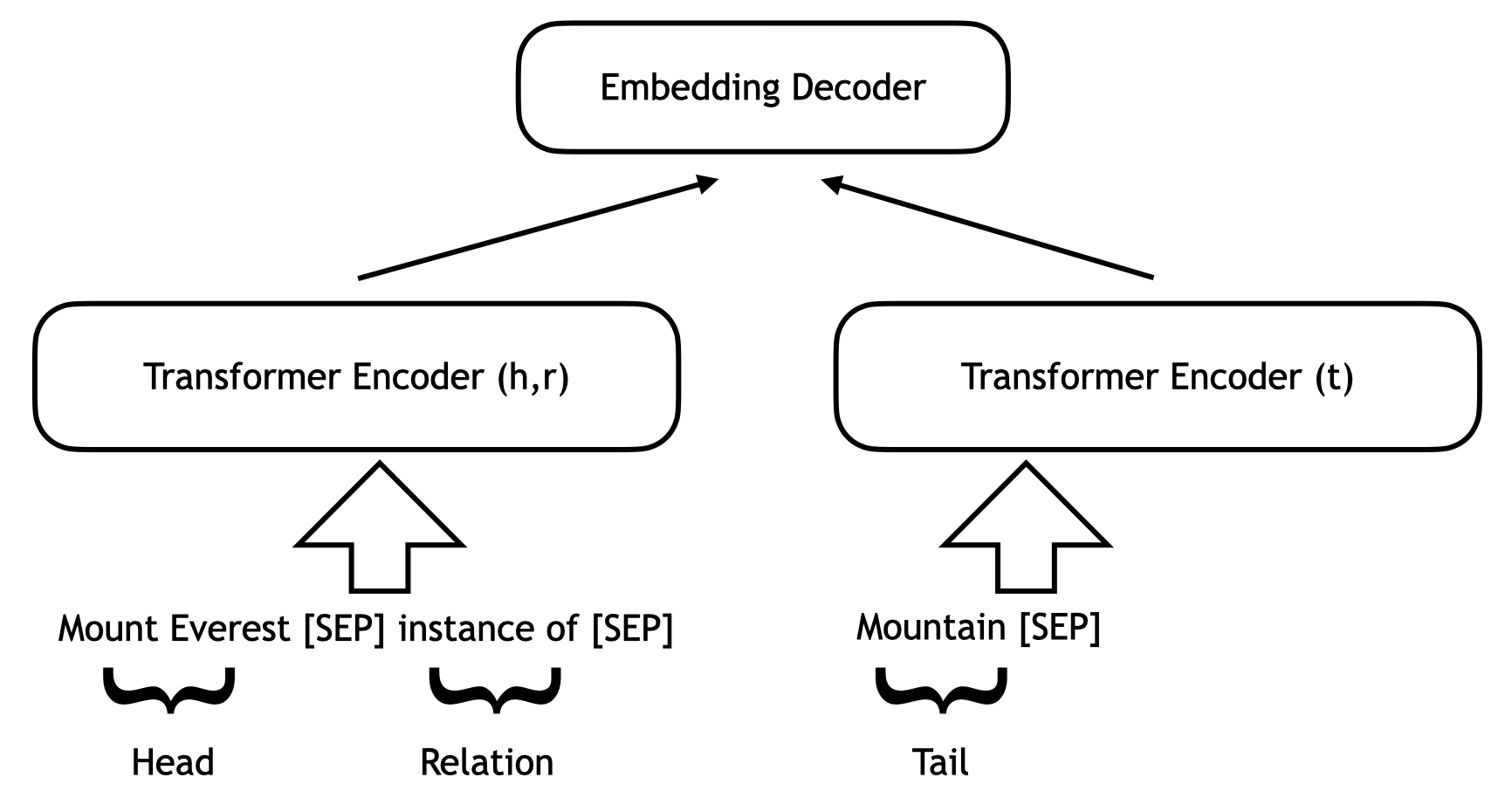}
\caption{An Illustrative example of the transformer-based approach for KG completion.}\label{fig:transformer_kgc}
\end{figure}

With the advent of large language models (LLM) in recent years, more and more NLP tasks are improved significantly improved by pretrained transformer-based models. In recent years, researchers have also started to think about using transformer-based models as solutions for KG-related tasks. A general illustration of the transformer-based approach for KG completion is shown in \ref{fig:transformer_kgc}. However, initial results from early papers have not fully demonstrated the effectiveness of language model-based solutions. It requires not only significantly more computational resources for training than KGE models, but also has slow inference speed. There are still many issues with PLM approach that are yet to be solved. 

The key difference between traditional KGE approach and PLM-based approach is that the former focuses on local structural information in graphs, whereas the latter relies on PLMs to decide the contextual relatedness between entities' names and descriptions. Textual descriptions of entities are usually available in many KGs such as Freebase, WordNet, and Wikidata. Triples are created from a large amount of corpus on entity description through information extraction. These entity descriptions are often stored in the knowledge base together with the entity entry. These textual descriptions are very useful information and one can use PLMs to extract textual features from these descriptions. PLMs are transformer-based models with many model parameters that are trained over large-scale corpora. PLMs are known to be able to generate good features for various NLP tasks. One of the famous PLMs is BERT \cite{devlin2018bert}. BERT is a pretrained bidirectional language model that is built based on transformer architecture. Since one of the tasks of BERT pretraining is next-sentence prediction, it naturally generates good features for characterizing whether 2 sentences are closely related. KG-BERT \cite{yao2019kg} is one of the first models we know of that uses PLMs to extract linguistic features from entity descriptions. It leverages the advantage of BERT, which is trained using next-sentence prediction to determine the association of head entity and tail entity descriptions for link prediction, and also triple classification tasks using the same logic.


ERNIE \cite{zhang2019ernie} proposes a transformer-based architecture that leverages lexical, syntactic, and knowledge information simultaneously by encoding textual description tokens through cascaded transformer blocks, as well as concatenating textual features and entity embedding features to achieve information fusion. K-BERT alleviates the knowledge noise issue by introducing soft positions and a visible matrix. StAR \cite{wang2021structure} and SimKGC \cite{wang2022simkgc} both use two separate transformer encoders to extract the textual representation of (head entity, relation) and (tail entity). However, they adopt very different methods to model the interaction between two encodings. StAR learns from previous NLP literature \cite{bowman2015large, reimers2019sentence} to apply interactive concatenation of features such as multiplication, subtraction, and the embedding vectors themselves to represent the semantic relationship between two parts of triples. On the other hand, SimKGC computes cosine similarity between two textual encodings. SimKGC also proposes new negative sampling methods including in-batch negatives, pre-batch negatives, and self negatives to improve the performance. Graph structural information is also considered in SimKGC to boost the score of entities that appear in the K-hop neighborhood. KEPLER \cite{wang2021kepler} uses the textual descriptions of head and tail entities as initialization for entity embeddings and uses TransE embedding as a decoder. The masked language modeling (MLM) loss is added to the knowledge embedding (KE) loss for overall optimization. InductivE \cite{wang2021inductive} uses features from pretrained BERT as graph embedding initialization for inductive learning on commonsense KGs (CKG). Experimental results on CKG show that fastText features can exhibit comparable performance as that from BERT. BERTRL \cite{zha2022inductive} fine-tunes PLM by using relation instances and possible reasoning paths in the local neighborhood as training samples. Relation paths in the local neighborhood are known to carry useful information for predicting direct relations between two entities \cite{lao2011random, xiong2017deeppath, shen2020modeling}. In BERTRL, each relation path is linearized to a sequence of tokens. The target triple and linearized relation paths are each fed to a pretrained BERT model to produce a likelihood score. The final link prediction decisions are made by aggregating these scores. The basic logic in this work is to perform link prediction through the relation path around the target triple. KGT5 \cite{saxena2022kgt5} propose to use a popular seq2seq transformer model named T5 \cite{raffel2020exploring} to pretrain on the link prediction task and perform KGQA. During link prediction training, a textual signal ``predict tail'' or ``predict head'' is prepended to the concatenated entity and relation sequence that's divided by a separation token. This sequence is fed to the encoder and the decoder's objective is to autoregressively predict the corresponding tail or head based on the textual signal. To perform question answering, textual signal ``predict answer'' is prepended to the query and we expect the decoder to autoregressively generate a corresponding answer to the query. This approach claims to significantly reduce the model size and inference time compared to other models. PKGC \cite{lv2022pre} proposes a new evaluation metric that is more accurate under an open-world assumption (OWA) setting.

\section{Conclusion}\label{sec:conclusion}

In conclusion, this paper has provided a comprehensive overview of the current state of research in KGE. We have explored the evolution of KGE models, with a particular focus on two main branches: distance-based methods and semantic matching-based methods. Through our analysis, we have uncovered intriguing connections among recently proposed models and identified a promising trend that combines geometric transformations to enhance the performance of existing KGE models. Moreover, this paper has curated valuable resources for KG research, including survey papers, open KGs, benchmarking datasets, and leaderboard results for link prediction. We have also delved into emerging directions that leverage neural network models, including graph neural networks and PLM, and highlighted how these approaches can be integrated with embedding-based models to achieve improved performance on diverse downstream tasks. In the rapidly evolving field of KG completion, this paper serves as a valuable reference for researchers and practitioners, offering insights into the past developments, current trends, and potential future directions. By providing a unified framework in the form of CompoundE and CompoundE3D, we aim to inspire further innovation in KGE methods and facilitate the construction of more accurate and comprehensive KGs. As the demand for knowledge-driven applications continues to grow, the pursuit of effective KGE models remains a pivotal area of research, and this paper lays the groundwork for future advancements in the field.

\section{Acknowledgments}
The authors acknowledge the Center for Advanced Research Computing (CARC) at the
University of Southern California for providing computing resources that
have contributed to the research results reported within this publication. URL: \url{https://carc.usc.edu}.

\bibliography{references}
\bibliographystyle{unsrt} 

\end{document}